\documentclass[twoside,11pt]{article}
\usepackage{jair, theapa, rawfonts}
\usepackage{url}
\usepackage{latexsym}
\usepackage{amsthm}
\usepackage{amssymb}
\usepackage{amsfonts}
\usepackage{amsmath}
\usepackage{multirow}
\usepackage{url}
\usepackage{graphicx}
\usepackage{mdwlist}
\usepackage{booktabs}
\usepackage{subfigure}
\usepackage{tabularx}
\usepackage{color}

\ShortHeadings{Bilingual Distributed Word Representations from Document-Aligned Data}
{Vuli\'{c} and Moens}
\firstpageno{1}

\begin{document}


\title{Bilingual Distributed Word Representations from Document-Aligned Comparable Data}


\author{\name Ivan Vuli\'{c} \email iv250@cam.ac.uk \\ \addr University of Cambridge \\ Department of Theoretical and Applied Linguistics \\
       9 West Road, CB3 9DP, Cambridge, UK \thanks{This work was done while the first author was a postdoctoral researcher at CS Dept., KU Leuven.}
         \AND
	  \name Marie-Francine Moens \email marie-francine.moens@cs.kuleuven.be \\ \addr KU Leuven \\ Department of Computer Science \\
       Celestijnenlaan 200A, 3001 Heverlee, Belgium
}


\maketitle

\begin{abstract}
We propose a new model for learning bilingual word representations from non-parallel document-aligned data. Following the recent advances in word representation learning, our model learns dense real-valued word vectors, that is, bilingual word embeddings (BWEs). Unlike prior work on inducing BWEs which heavily relied on parallel sentence-aligned corpora and/or readily available translation resources such as dictionaries, the article reveals that BWEs may be learned solely on the basis of document-aligned comparable data without any additional lexical resources nor syntactic information. We present a comparison of our approach with previous state-of-the-art models for learning bilingual word representations from comparable data that rely on the framework of multilingual probabilistic topic modeling (MuPTM), as well as with distributional local context-counting models. We demonstrate the utility of the induced BWEs in two semantic tasks: (1) bilingual lexicon extraction, (2) suggesting word translations in context for polysemous words. Our simple yet effective BWE-based models significantly outperform the MuPTM-based and context-counting representation models from comparable data as well as prior BWE-based models, and acquire the best reported results on both tasks for all three tested language pairs.
\end{abstract}

\section{Introduction}
\label{s:intro}
%

A huge body of work in distributional semantics and word representation learning almost exclusively revolves around the {\em distributional hypothesis} \cite{Harris:1954} - an idea which states that similar words occur in similar contexts. All current corpus-based approaches to semantics rely on the contextual evidence in one way or another. Roughly speaking, word representations are typically learned using these two families of distributional context-based models: (1) global matrix factorization models such as latent semantic analysis (LSA) \cite{Landauer:19971} or generative probabilistic models such as latent Dirichlet allocation (LDA) \cite{Blei:2003}, which model the word co-occurrence at the document or paragraph level; or (2) local context window models that represent words as sparse high-dimensional context vectors, and model the word co-occurrence at the level of selected neighboring words \cite{Turney:2010}, or generative probabilistic models that learn the probability distribution of a vocabulary word in the context window as a latent variable \cite{Deschacht:2009,Deschacht:2012}.

On the other hand, dense real-valued vectors known as distributed representations of words or {\em word embeddings} (WEs) \cite<e.g.,>{Bengio:2003,Collobert:2008,Mikolov:2013iclr,Pennington:2014} have been introduced recently, first as part of neural network based architectures for statistical language modeling. 
WEs serve as richer and more coherent word representations than the ones obtained by the aforementioned traditional distributional semantic models, with illustrative comparative studies available in the recently published relevant work \cite<e.g.,>{Mikolov:2013naacl,Baroni:2014,Levy:2015tacl}. 

A natural extension of interest from monolingual to multilingual word embeddings has occurred recently \cite<e.g.,>{Klementiev:2012,Hermann:2014}. When operating in multilingual settings, it is highly desirable to learn embeddings for words denoting similar concepts that are very close in the {\em shared bilingual embedding space} (e.g., the representations for the English word {\em school} and the Spanish word {\em escuela} should be very similar). These BWEs may then be used in a myriad of multilingual natural language processing tasks and beyond, such as fundamental tasks leaning on such bilingual meaning representations, e.g., computing cross-lingual and multilingual semantic word similarity and extracting bilingual word lexicons using the induced bilingual embedding space (see Figure~\ref{fig:monomulti}). However, all these models critically require (at least) sentence-aligned parallel data and readily-available translation dictionaries to induce {\em bilingual word embeddings} (BWEs) that are consistent and closely aligned over different languages.

\noindent {\bf Contributions.} To the best of our knowledge, this article presents the first work to showcase that bilingual word embeddings may be induced directly on the basis of comparable data without any additional bilingual resources such as sentence-aligned parallel data or translation dictionaries. The focus is on document-aligned comparable corpora (e.g., Wikipedia articles aligned through inter-wiki links, news texts discussing the same theme).  

Our new bilingual distributed representation learning model makes use of {\em pseudo-bilingual documents} constructed by merging the content of two coupled documents from a document pair, where we propose and evaluate two different strategies on how to construct such pseudo-bilingual documents: (1) {\em merge and randomly shuffle} strategy which randomly permutes words from both languages in each pseudo-bilingual document, and (2) {\em length-ratio shuffle} strategy, a deterministic method that retains monolingual word order while intermingling the words cross-lingually. These additional pre-training shuffling strategies ensure that both source language words and target language words occur in the contexts of each source and target language word. A monolingual model such as {\em skip-gram with negative sampling (SGNS)} from the \texttt{word2vec} package \cite{Mikolov:2013nips} is then trained on these ``shuffled'' pseudo-bilingual documents. By this procedure, we steer semantically similar words from different languages towards similar representations in the shared bilingual embedding space, and effectively use available bilingual contexts instead of monolingual ones. The model treats documents as bags-of-words (i.e., it does not include any syntactic information) and does not even rely on any sentence boundary information. 

In summary, the main contributions of this article are: \\
\\
\noindent {\bf (1)} We present BWE Skip-Gram (BWESG), the first model that induces bilingual word embeddings directly from document-aligned non-parallel data. We test and evaluate two main variants of the model based on the pre-training shuffling step. The main strength of the presented model lies in its favourable trade-off between simplicity and effectiveness. \\
\noindent {\bf (2)} We provide a qualitative and quantitative analysis of the model. We draw analogies and comparisons with prior work on inducing word representations from the same data type: document-aligned comparable corpora (e.g., models relying on the multilingual probabilistic topic modeling framework (MuPTM)). \\
\noindent {\bf (3)} We demonstrate the utility of induced BWEs {\em at the word type level} in the task of bilingual lexicon extraction (BLE) from Wikipedia data for three language pairs. A BLE model based on our BWEs significantly outperforms MuPTM-based and context-counting BLE models, and acquires the best reported scores on the benchmarking BLE datasets. \\
\noindent {\bf (4)} We demonstrate the utility of induced BWEs {\em at the word token level} in the task of suggesting word translations in context (SWTC) \cite{Vulic:2014emnlp} for the same three language pairs. A SWTC model based on our BWEs again significantly outscores the best scoring MuPTM-based SWTC models in the same setting without any use of parallel data and translation dictionaries, and again acquires the best reported results on the benchmarking SWTC datasets. \\
\noindent {\bf (5)} We also present a comparison with state-of-the-art BWE induction models \cite{Mikolov:2013arxiv,Hermann:2014,Gouws:2015icml} in BLE and SWTC. Results reveal that our simple yet effective approach is on-par with or outperforms other BWE induction models that rely on parallel data or readily available dictionaries to learn shared bilingual embedding spaces. In addition, preliminary experiments with BWESG on parallel Europarl data demonstrate that the model is also useful when trained on sentence-aligned data, reaching the performance of benchmarking BWE induction models from parallel data \cite<e.g.,>{Hermann:2014}. 
\vspace{-0.7em}

\section{Related Work}
\label{s:related}
In this section we further motivate why we opt for building a model for inducing bilingual word embeddings from comparable document-aligned data. For a clearer overview, we have split related work into three broad clusters: (1) monolingual word embeddings, (2) bilingual word embeddings, and (3) bilingual word representations from document-aligned data.
\vspace{-0.0em}
\subsection{Monolingual Word Embeddings}
\label{ss:rwmwe}
The idea of representing words as continuous real-valued vectors dates way back to mid-80s \cite{Rumelhart:1986,Elman:1990}. The idea met its resurgence a decade ago \cite{Bengio:2003}, where a neural language model learns word embeddings as part of a neural network architecture for statistical language modeling. This work inspired other approaches that learn word embeddings within the neural-network language modeling framework \cite{Collobert:2008,Collobert:2011}. Word embeddings are tailored to capture semantics and encode a continuous notion of semantic similarity (as opposed to semantically poorer discrete representations), necessary to share information between words and other text units.
\begin{figure}[!t]
\begin{center}
\includegraphics[width=0.85\textwidth]{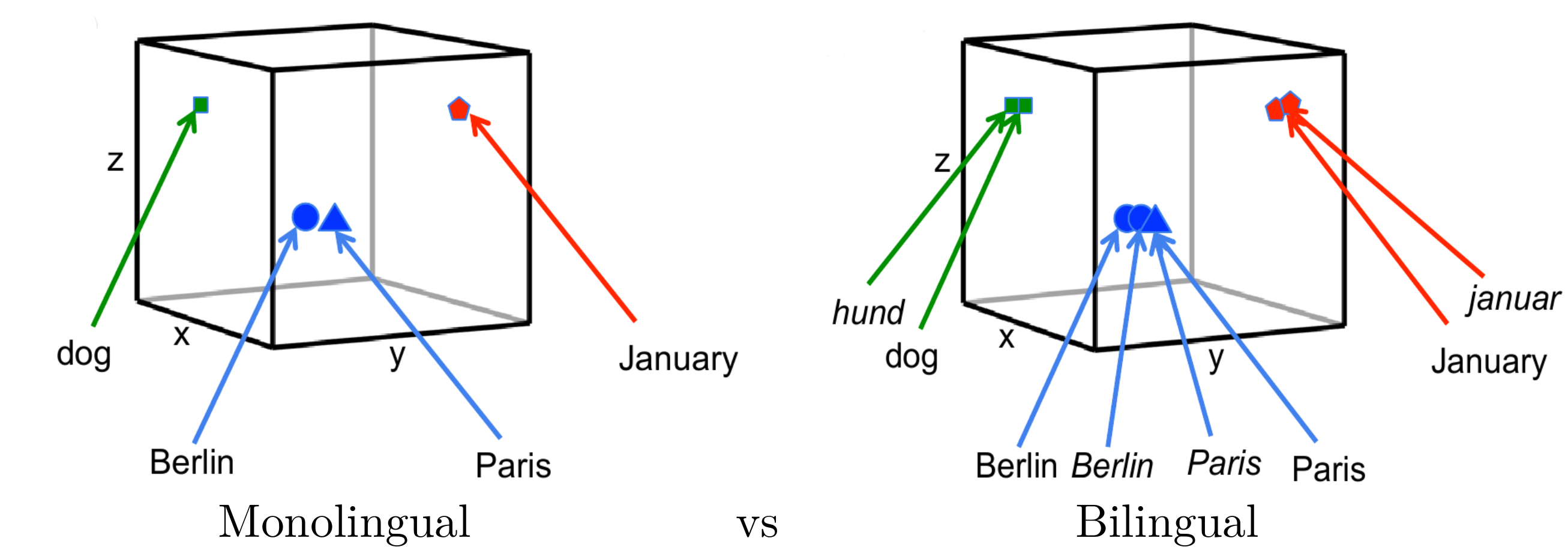}
\vspace{-0.7em}
\caption{A toy 3D shared bilingual embedding space from \citeA{Gouws:2015icml}: While in monolingual spaces words with similar meanings should have similar representations, in bilingual spaces words in two different languages with similar meanings should have similar representations (both mono- and cross-lingually).}
\label{fig:monomulti}
\vspace{-2.5em}
\end{center}
\end{figure}

Recently, the skip-gram and continuous bag-of-words (CBOW) model from Mikolov et al. \citeyear{Mikolov:2013iclr,Mikolov:2013nips} revealed that the full neural-network structure is not needed at all to learn high-quality word embeddings (with extremely decreased training times compared to the full-fledged neural network models, see \citeS{Mikolov:2013iclr} work for the full analysis of complexity of the models). These models are in fact simple single-layered architectures, where the objective is to predict a word's context given the word itself (skip-gram) or predict a word given its context (CBOW). Similar models called vector log-bilinear models were recently proposed \cite{Mnih:2013}. Other models inspired by skip-gram and CBOW are GloVe (Global Vectors for Word Representation) \cite{Pennington:2014}, which combines local and global contexts of a word into a unified model, and a model which relies on dependency-based contexts instead of simpler word-based contexts \cite{Levy:2014acl}, and new models are steadily emerging \cite<e.g.,>{Lebret:2014eacl,Lu:2015naacl,Stratos:2015acl,Trask:2015icml,Liu:2015acl}. 

An interesting finding has been discussed recently \cite{Levy:2014}: the popular skip-gram model with negative sampling (SGNS) \cite{Goldberg:2014arxiv} is simply a model which implicitly factorizes a word-context matrix, with its cells containing pointwise mutual information (PMI) scores of the respective word and context pairs, shifted by a global constant. In other words, the SGNS performs exactly the same thing as traditional distributional models (i.e., context counting plus context weighting and/or dimensionality reduction), with a slight improvement in performance with SGNS \cite{Baroni:2014,Levy:2015tacl}. 

All these low-dimensional vectors, besides improving computational efficiency, lead to better generalizations, even allowing to generalize over the vocabularies observed in labelled data, and hence partially alleviating the ubiquitous problem of data sparsity. Their utility has been validated and proven in various semantic tasks such as semantic word similarity, synonymy detection or word analogy solving \cite{Mikolov:2013naacl,Baroni:2014,Pennington:2014}. Moreover, word embeddings have been proven to serve as useful unsupervised features for plenty of downstream NLP tasks such as named entity recognition, chunking, semantic role labeling, part-of-speech tagging, parsing, selectional preferences \cite{Turian:2010,Collobert:2011,Chen:2014emnlp}.

Due to its simplicity, as well as its efficacy and consequent popularity in various tasks \cite{Mikolov:2013nips,Levy:2014}, with a clear advantage on similarity tasks when compared to traditional models from distributional semantics \cite{Levy:2015tacl} in this article we will focus on the adaptation of SGNS \cite{Mikolov:2013nips}. In Section~\ref{s:architecture}, we provide a very brief overview of the model, and then follow up with our new bilingual model which is based on SGNS.

\subsection{Bilingual Word Embeddings} 
\label{ss:rwbwe}
Bilingual word representations could serve as an useful source knowledge for problems in cross-lingual information retrieval \cite{Levow:2005,Vulic:20121}, statistical machine translation \cite{WuWang:2008}, document classification \cite{Ni:2011,Klementiev:2012,Hermann:2014,Chandar:2014,Vulic:2015ipm}, bilingual lexicon extraction \cite{Tamura:2012,Vulic:2013naacl}, or knowledge transfer and annotation projection from resource-rich to resource-poor languages for a myriad of NLP tasks such as dependency parsing, POS tagging, semantic role labeling or selectional preferences \cite{Yarowsky:2001,Pado:2009,Peirsman:2010,Das:2011,Tackstrom:2013tacl,Ganchev:2013,Tiedemann:2014,Xiao:2014}. Other interesting application domains are machine translation \cite<e.g.,>{Zou:2013,Wu:2014,Zhang:2014} and cross-lingual information retrieval \cite<e.g.,>{Vulic:2015sigir}. Moreover, by making the transition from monolingual to bilingual settings and building a {\em shared bilingual embedding space} (see again Figure~\ref{fig:monomulti} for an illustrative example), one is able to extend or rather generalize semantic tasks such as semantic similarity computation, synonymy detection or word analogy computation across languages. Following the success in monolingual settings, a body of recent work on word representation learning has therefore focused on learning bilingual word embeddings (BWEs).

The current research on inducing BWEs critically relies on sentence-aligned parallel data or readily available bilingual lexicons to achieve the coherence of representations across languages (e.g., to build similar representations for similar concepts in different languages such as {\em January-januari}, {\em dog-hund} or {\em sky-hemel}). We may cluster the current work in three different groups: (1) the models that rely on hard word alignments obtained from parallel data to constrain the learning of BWEs \cite{Klementiev:2012,Zou:2013,Wu:2014}; (2) the models that use the alignment of parallel data at the sentence level \cite{Kocisky:2014,Hermann:2014iclr,Hermann:2014,Chandar:2014,Shi:2015acl,Gouws:2015icml}; (3) the models that critically require readily available bilingual lexicons \cite{Mikolov:2013arxiv,Faruqui:2014,Xiao:2014}. The main disadvantage of all these models is the limited availability of parallel data and bilingual lexicons, resources which are scarce and/or domain-restricted for plenty of language pairs. In this work, we significantly alleviate the requirements: unlike prior work, we show that BWEs may be induced solely on the basis of document-aligned comparable data without any additional need for parallel data or bilingual lexicons. Note that (in theory) the work from \citeA{Hermann:2014,Chandar:2014} may also be extended to the same setting with document-aligned data, as these two models originally rely on sentence embeddings computed as aggregations over their single word embeddings plus sentence alignments. In this work, by testing and comparing to the BiCVM model from \citeA{Hermann:2014}, we show that these models do not work well in practice after replacing the very strong bilingual signal coded in parallel sentences with the noisy bilingual signal given by document alignments and non-parallel data. 

\subsection{Bilingual Word Representations from Document-Aligned Data}
\label{ss:rwwr}
Prior work on inducing bilingual word representations in the early days followed the tradition of window-based context-counting distributional models \cite{Rapp:1999,Gaussier:2004,Laroche:2010} and it again required a bilingual lexicon as a critical resource. In order to tackle this issue, recent work relies on the supervision-lighter framework of multilingual probabilistic topic modeling (MuPTM) \cite{Mimno:2009,Boyd:2009,DeSmet:2009,Xiaochuan:2009,Zhang:2010,Fukumasu:2012} or other similar models for latent structure induction \cite{Haghighi:2008,Daume:2011}. 

Words in this setting are represented as real-valued vectors with conditional topic probability scores $P(z_k|w_i)$, regardless of their actual language. Topics $z_k$ are in fact latent inter-lingual concepts discovered directly from multilingual comparable data using a multilingual topic model such as bilingual LDA. We discuss the MuPTM-based representations in more detail in Section~\ref{ss:brm}. 

MuPTM-based bilingual word representations induced from comparable data have demonstrated its utility in tasks such as cross-lingual semantic similarity computation and bilingual lexicon extraction \cite{Vulic:2011,Liu:2013} and suggesting word translations in context \cite{Vulic:2014emnlp}. In this work, we compare the state-of-the-art MuPTM-based word representations induced from the same type of comparable corpora with BWEs learned by our new model in these two semantic tasks.

Another recent model \cite{Sogaard:2015acl} is also able to learn from document-aligned data. It is a count-based model which builds binary word vectors denoting the occurrence of each word in each document pair. Dimensionality reduction is then applied post-hoc on the induced sparse vectors. Since the links between documents are known, the model is able to learn cross-lingual correspondences between words and, consequently, bilingual word representations. Exactly the same idea was already introduced as a baseline model by \citeA{Vulic:2011}, where TF-IDF weights were used instead of binary indices, and no dimensionality reduction was applied post-hoc. The model from \citeA{Vulic:2011} was surpassed by baseline models from document-aligned data briefly discussed in Section~\ref{ss:brm}, while the model from \citeA{Sogaard:2015acl} obtains results that are very similar to the BWE baselines compared against in this work (described in Section~\ref{ss:brm2}).
\section{BWESG: Model Architecture}
\label{s:architecture}
Our new bilingual model is an extension of SGNS to bilingual settings with document-aligned comparable training data. This section describes the underlying SGNS and two variants of our SGNS-based BWE induction model. 


\subsection{Skip-Gram with Negative Sampling (SGNS)}
\label{ss:archsg}
Our departure point is the log-linear SGNS from \citeA{Mikolov:2013nips} as implemented in the \texttt{word2vec} package.\footnote{\texttt{https://code.google.com/p/word2vec/}} The SGNS model learns word embeddings (WEs) in a similar way to neural language models \cite{Bengio:2003,Collobert:2008}, but without a non-linear hidden layer.

In the monolingual setting, we assume one language $L$ with vocabulary $V$, and a corpus of words $w \in V$, along with their contexts $c \in V^c$, where $V^c$ is the context vocabulary. Contexts for each word $w_n$ are typically neighboring words in a context window of size $cs$ (i.e., $w_{n-cs},\ldots,w_{n-1},w_{n+1},\ldots,w_{n+cs}$), so effectively it holds $V^c \equiv V$.\footnote{Testing other options for context selection such as dependency-based contexts \cite{Levy:2014acl} is beyond the scope of this work, and it was shown that these contexts may not lead to any gains in the final WEs \cite{Kiela:2014}.}

Each word type $w \in V$ is associated with a vector $\vec{w} \in \mathbb{R}^d$ (its pivot word representation or pivot word embedding, see Figure~\ref{fig:model}), and a vector $\vec{w_c} \in \mathbb{R}^d$ (its context embedding). $d$ is the dimensionality of the WE vectors, which, as a model input parameter, has to be set in advance before the training procedure commences. The entries in these vectors are latent, and treated as parameters $\theta$ to be learned by the model. In short, the idea of the skip-gram model is to scan through the corpus (which is typically unannotated, \citeR{Mikolov:2013iclr}) {\em word by word} in turn (i.e., these are the pivot words), and learn from the pairs {\em (word, context word)}. The learning goal is to maximize the ability of predicting context words for each pivot word in the corpus. Let $ob=1$ denote that the pair of words $(w,v)$ is observed in the corpus and thus belongs to the training set $D$. The probability of $(w,v) \in D$ is defined by the softmax function:
\begin{align}
P(ob=1|w,v,\theta)=\frac{1}{1+\exp(-\vec{w} \cdot \vec{v_c})}
\end{align}
Each word token $w$ in the corpus is treated in turn as the pivot and all pairs of word tokens ($w,w\pm 1$),...,($w,w\pm t(cs)$) are appended to $D$, where $t(cs)$ is an integer sampled from a uniform distribution on $\{1,\ldots,cs\}$.\footnote{The original skip-gram model utilizes dynamic window sizes, where $cs$ denotes the maximum window size. Moreover, the model takes into account sentence boundaries in context selection, that is, it selects as context words only words occurring in the same sentence as the pivot word.} The global training objective $J$ is then to maximize the probabilities that all pairs from $D$ are indeed observed in the corpus:
\begin{align}
J=\arg \max_{\theta} \sum_{(w,v) \in D} \log \frac{1}{1+\exp(-\vec{w} \cdot \vec{v_c})}
\end{align}
where $\theta$ are the parameters of the model, that is, pivot and context word embeddings which have to be learned. One may see that this objective function has a trivial solution by setting $\vec{w}=\vec{v_c}$, and $\vec{w}\cdot\vec{v_c}=Val$, where $Val$ is a large enough number \cite{Goldberg:2014arxiv}. In order to prevent this trivial training scenario, the {\em negative sampling} procedure comes into the picture \cite{Collobert:2008,Mikolov:2013nips}. 

In short, the idea behind negative sampling is to present the model with a set $D'$ of artificially created or sampled ``negative pivot-context'' word pairs ($w,v'$), which by assumption serve as negative examples, that is, they do not occur as observed/positive {\em (word, context)} pairs in the training corpus. The model then has to adjust the parameters $\theta$ in such a way to also maximize the probability that these negative pairs will not occur in the corpus. While the interested reader may find further details about the negative sampling procedure, and the new exact objective function along with its derivation elsewhere \cite{Levy:2014}, for illustrative purposes and simplicity, here we present the approximative objective function with negative sampling by Goldberg and Levy \citeyear{Goldberg:2014arxiv}: 
\begin{align}
J=\arg \max_{\theta} &\sum_{(w,v) \in D} \log \frac{1}{1+\exp(-\vec{w} \cdot \vec{v_c})} + \sum_{(w,v') \in D'} \log \frac{1}{1+\exp(\vec{w} \cdot \vec{v'_c})}
\label{eq:objective}
\end{align}
The free parameters $\theta$ are updated using stochastic gradient descent and backpropagation, with learning rate typically controlled by Adagrad \cite{Duchi:2011} or with a global linearly decreasing learning rate. By optimizing the objective from eq.~\eqref{eq:objective}, the model incrementally pushes observed pivot WEs towards context WEs of their collocates in the corpus. In the words of distributional hypothesis - after training, words that occur in similar contexts should end up having similar word embeddings. In other words, to link the terminology of distributional hypothesis and the modeling assumptions of SGNS - words that predict similar contexts end up having similar word embeddings.
\subsection{Final Model - BWESG: BWE Skip-Gram} 
\begin{figure}[t]
						\centering						
						\subfigure[Merge and Shuffle]{						
							\includegraphics[width=0.98\textwidth]{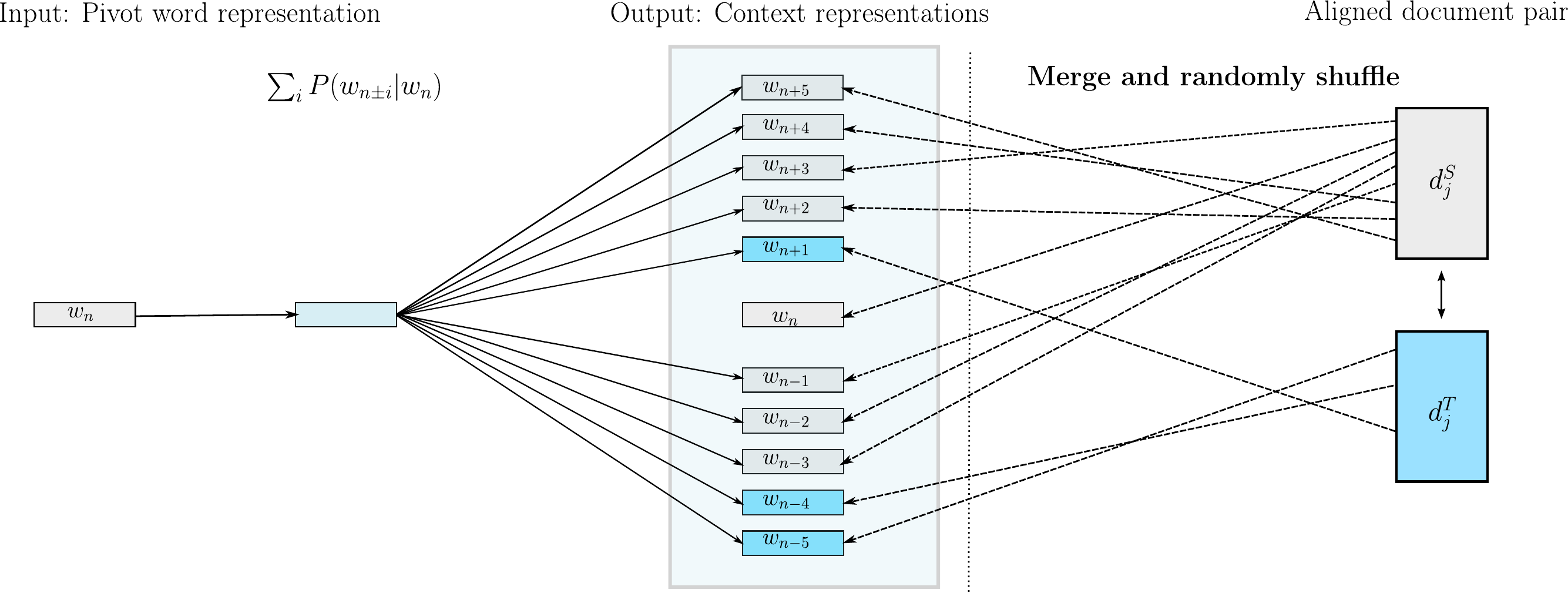}
							\label{fig:random}
						}
						\subfigure[Length-Ratio Shuffle]{
							\includegraphics[width=0.98\textwidth]{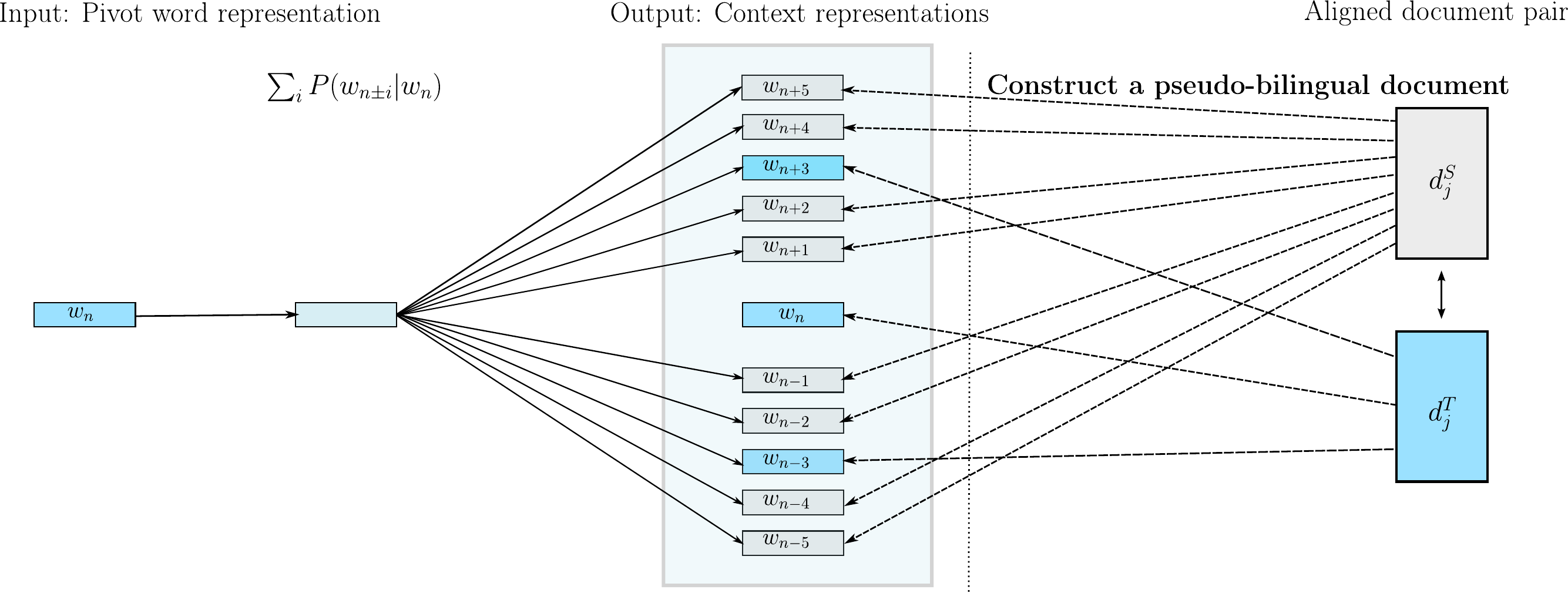}	
							\label{fig:nonrandom}				
						}
						\vspace{-0.7em}
						\caption{The architecture of our BWE Skip-Gram (BWESG) model for learning bilingual word embeddings from document-aligned comparable data with two different pre-training strategies: (1) non-deterministic {\em merge and shuffle}, (2) deterministic {\em length-ratio shuffle}. Source language words and documents are drawn as gray boxes, while target language words and documents are drawn as blue boxes. The right side of the figures (separated by vertical dashed lines) illustrates how a pseudo-bilingual document is constructed from a pair of two aligned documents.}
					\label{fig:model}
					\vspace{-1.8em}
\end{figure}

In the next step, we propose a novel method that extends SGNS to work with bilingual document-aligned comparable data. 
Let us assume that we possess a document-aligned comparable corpus, defined as $\mathcal{C} = \{d_1,d_2,\ldots,d_N\}=\{(d_1^S,d_1^T),(d_2^S,d_2^T),\allowbreak \ldots,(d_N^S,d_N^T)\}$. $d_j=(d_j^S,d_j^T)$ denotes a pair of aligned documents in the source language $L_S$ and the target language $L_T$ respectively, and $N$ is the number of pairs in the corpus. $V^S$ and $V^T$ are vocabularies associated with languages $L_S$ and $L_T$. The goal is to learn a shared bilingual embedding space given the data (Figure~\ref{fig:monomulti}) and document alignments as the only bilingual signal during training. We present two strategies that, coupled with SGNS, lead to such shared bilingual spaces. An overview of the architecture for learning BWEs from document-aligned comparable data with the two strategies is given in Figures~\ref{fig:random} and \ref{fig:nonrandom}. \\
\\
\noindent {\bf (1) Merge and Shuffle.} In the first step, we {\em merge} two documents $d_j^S$ and $d_j^T$ from the aligned document pair $d_j$ into a single ``pseudo-bilingual'' document $d'_j$. Following that, we randomly {\em shuffle} the newly constructed pseudo-bilingual document. A shuffle is a (random) permutation of the word tokens given in two different languages forming the pseudo-bilingual document. The pre-training shuffling step (see Figure~\ref{fig:random}) assures that each word $w$, regardless of its actual language, obtains word collocates from both vocabularies. The idea of obtaining bilingual contexts for each pivot word in each pseudo-bilingual document will steer the final model towards constructing a shared bilingual space. Since the model depends on the alignment at the document level, in order to ensure the bilingual contexts instead of monolingual contexts, it is intuitive to assume that larger window sizes will lead to better bilingual embeddings. We test this hypothesis and the effect of window size in Section~\ref{ss:blerd}. In another interpretation, since the model relies only on (pseudo-bilingual) document level co-occurrence, the window size parameter then just controls the amount of random data dropout, that is, the number of positive document-level training examples. The locality feature of SGNS is not preserved due to the shuffling procedure. \\
\\
\noindent {\bf (2) Length-Ratio Shuffle.} The non-deterministic and uncontrollable nature of the {\em merge and shuffle} procedure opens up a possibility of accidentally obtaining ``bad shuffles'' that will result in sub-optimal word representations. Therefore, we also propose a {\em deterministic strategy} for building pseudo-bilingual documents suitable for bilingual training. Source and target language words are inserted into an (initially empty) pseudo-bilingual document in turn based on the ratio of document lengths, with word order preserved. Document lengths are measured in terms of word tokens, and let us denote them as $m_S$ and $m_T$ for an aligned document pair $(d_j^S,d_j^T)$. Let us assume, without loss of generality, that $m_S \geq m_T$. The procedure then proceeds as follows (if $m_T > m_S$ the procedure proceeds in an analogous manner with the roles of $d_j^S$ and $d_j^T$ reversed): 
\begin{enumerate*}
\item Pseudo-bilingual document $d'_j$ is empty: $d'_j=\{\}$.
\item Compute the ratio: $R=\lfloor \frac{m_S}{m_T} \rfloor$.
\item Scan through aligned documents $d_S$ and $d_T$ simultaneously and (3.1) append $R$ word tokens from $d_j^S$ into $d'_j$; then (3.2) append 1 word token from $d_j^T$. Repeat steps 3.1 and 3.2 until all word tokens from $d_j^T$ have been inserted into $d'_j$.
\item Insert remaining $m_S \mod m_T$ word tokens from $d_j^S$ into $d'_j$.
\end{enumerate*}
Using a simple example, assume that we have an English (EN) document $\{Frodo, Sam, orcs,\break goblins, Mordor, ring\}$ and a Spanish (ES) document $\{anillo, orcos, mago\}$: the pseudo-bilingual document would be formed by inserting 1 Spanish word after 2 English words (as the length ratio is 6:3 = 2:1). The final pseudo-bilingual document is: \\
{\small
$\{Frodo_{EN}, Sam_{EN}, anillo_{ES}, orcs_{EN}, goblins_{EN}, orcos_{ES}, Mordor_{EN}, ring_{EN}, {mago}_{ES}\}$.}

In another interpretation, the {\em length-ratio shuffle} strategy constructs a single permutation/shuffle of the pseudo-bilingual document controlled by the word order in two aligned documents as well as their length ratio. As before, the model relies on pseudo-bilingual document level co-occurrence, and the window size parameter controls the amount of (now non-random)
data dropout. A difference lies in the fact that this procedure now keeps word order intact monolingually while constructing a pseudo-bilingual document.

The final BWE Skip-gram (BWESG) model then relies on the monolingual variant of SGNS (or any other monolingual WE induction model) trained on these shuffled/permuted pseudo-bilingual documents (using strategies (1) or (2)).\footnote{We were also experimenting with GloVe and CBOW, but they were falling short of SGNS on average.} The model learns word embeddings for source and target language words aligned over the $d$ shared embedding dimensions. The BWESG-based representation of word $w$, regardless of its actual language, is then a $d$-dimensional vector: $\vec{w}=[f_1,\ldots, f_k,\ldots, f_d]$. $f_k \in \mathbb R$ denotes the score for the $k$-th shared inter-lingual feature within the $d$-dimensional shared bilingual embedding space. Since all words share the embedding space, semantic similarity between words may be computed both monolingually and across languages. We will extensively use this property in our evaluation tasks.
\section{Baseline Representation Models}
\label{s:baselinemodels}
We quickly navigate through other approaches to bilingual word representation learning from document-aligned comparable data. The set of models in comparison may be roughly clustered into two main groups: (Group I) ``pre-BWE'' baseline representation models from document-aligned data, (Group II) benchmarking BWE induction models that were not originally developed for learning from document-aligned comparable data. While it is essential to compare the BWESG model with other frameworks for learning representations from document-aligned data (Group I), it is also crucial to detect main strengths of the BWESG model when compared to other approaches in the BWE learning framework which can also be adjusted to learn from document-aligned data (Group II). 

\subsection{Group I: Baseline Representation Models from Document-Aligned Data}
\label{ss:brm}

\noindent {\bf Basic-MuPTM} The early approaches \cite<e.g.,>{Dumais:1996,Carbonell:1998} tried to mine topical structure from document-aligned comparable texts using a monolingual topic model (e.g., LSA or LDA) trained on pseudo-bilingual documents with the target document simply appended to its source language counterpart, and then used the discovered latent topical structure as a shared semantic space in which both words and documents from two languages may be represented in a uniform way. 

More recent work on multilingual probabilistic topic modeling (MuPTM) \cite{Mimno:2009,DeSmet:2009,Vulic:2011} showed that word representations of higher quality may be built if a multilingual topic model such as bilingual LDA (BiLDA) is trained jointly on document-aligned comparable corpora by retaining the structure of the corpus intact (i.e., there is no need to construct pseudo-bilingual documents).

MuPTM discovers the latent structure of the observed data in the form of $K$ latent cross-lingual topics $z_1,\ldots,z_K$ which optimally describe the generation of observed data. Extracting latent cross-lingual topics actually implies learning per-document topic distributions for each document in the corpus (probability scores $P(z_k|d_j)$), and discovering language-specific representations of these topics given by per-topic word distributions in each language (probability scores $P(w_i^S|z_k)$ and $P(w_i^T|z_k)$). Latent cross-lingual topics are in fact distributions over vocabulary words, and have their language-specific representation in each language. Per-document topic distributions and per-topic word distributions are obtained after training the topic model on multilingual data. The representation of some word $w \in V^S$ (or in an analogous manner $w \in V^T$) is then a $K$-dimensional vector: \\
$\vec{w}=[P(z_1|w),\ldots,P(z_k|w),\ldots,P(z_K|w)]$. 

We call this representation model (RM) {\em Basic-MuPTM (BMu)}. Since the number of topics, that is, the number of vector dimensions $K$ is typically high \cite{Dinu:2010,Vulic:2011}, additional feature pruning \cite{Reisinger:2010} may be employed in order to retain only the most descriptive dimensions in the MuPTM-based representation, which was shown to improve the performance on several semantic tasks (e.g., BLE or SWTC) \cite{Vulic:2013naacl,Vulic:2015ipm}. 

A multilingual topic model is typically trained by Gibbs sampling \cite{Geman:1984,Steyvers:2007,Vulic:2015ipm}. Similar to the SGNS/BWESG training procedure, Gibbs sampling for MuPTM/BiLDA also scans the training corpus word by word, and then cyclically updates topic assignments for each word token. However, unlike BWESG which uses only a subset of document-level training examples, Gibbs sampling for MuPTM uses all words from the source language document as well as all words from its coupled target language document to influence the topic assignment for the pivot word. The BWESG design relying on data dropout leads to decreased training times and computation costs to obtain final representations compared to Basic-MuPTM. \\ 
\\
\noindent {\bf Association-MuPTM} Another representation is also based on the MuPTM framework: it contains association scores $P(w_a|w)$ for each $w,w_a \allowbreak \in V^S \cup V^T$ \cite{Vulic:2013naacl} as dimensions of real-valued word vectors. These association scores are computed as $P(w_a|w)=\allowbreak \sum_{k=1}^K \allowbreak P(w_a|z_k) \allowbreak P(z_k|w)$ \cite{Griffiths:2007}, and the word vector is a ($|V^S|+|V^T|$)-dimensional vector: $\vec{w}=[P(w_1^S|w),\ldots,P(w^S_{|V^S|}|w),P(w_1^T|w),\ldots,P(w^T_{|V^T|}|w)]$. As with Basic-MuPTM, the original word representation may also be pruned post-hoc. We call this representation model {\em Association-MuPTM (AMu)}. Since this approach relies on the MuPTM training plus additional $|V^S|\cdot|V^T|$ computations to estimate association scores, the cost of obtaining Association-MuPTM representations is even higher than for Basic-MuPTM, but it leads to more robust word representations for the BLE task \cite{Vulic:2013naacl}. While both Basic-MuPTM and Association-MuPTM produce high-dimensional real-valued vectors with plenty of near-zero dimensions (the number of dimensions is typically measured in thousands) which have to be pruned afterwards with the pruning parameter often set ad-hoc, BWESG produces lower-dimensional dense real-valued vectors, and no additional post-hoc feature pruning is required for BWESG. \\
\\
\noindent {\bf Traditional-PPMI} A traditional approach to building bilingual word representations in (cross-lingual) distributional semantics is to compute weighted co-occurrence scores (e.g., using PMI, TF-IDF) between pivot words and their context words in a window of predefined size, plus an external bilingual lexicon to align context words/dimensions across languages \cite{Gaussier:2004,Laroche:2010}. A weighting function (WeF), which is a standard choice in distributional semantics and yields optimal or near-optimal results over a group of semantic tasks \cite{Bullinaria:2007}, is the smoothed positive pointwise mutual information statistic \cite{Pantel:2002,Turney:2010}. Furthermore, in order to induce context words without the need for a readily available lexicon, we employ the bootstrapping procedure from \citeA{Peirsman:2011,Vulic:2013emnlp}. This representation model is called {\em Traditional-PPMI (TPPMI)}. The word representation is an $R$-dimensional vector: $\vec{w}=[sc_1(w,c_1),\ldots, sc_k(w,c_k),\ldots, sc_R(w,c_R)]$. The dimensions of the vector space are $R$ one-to-one word translation pairs $c_k=(c_k^S,c_k^T)$, and $sc_k(w,c_k)$ is the weighted co-occurrence score of the pivot word $w$ and the $k$-th context feature, where one computes the co-occurrence score using $c_k^S$ if $w \in V^S$, or $c_k^T$ if $w \in V^T$.

Vector dimensions $c_k=(c_k^S,c_k^T)$ in the Traditional-PPMI representation and similar models with other WeFs are typically the most frequent and reliable translation pairs in the corpus. As opposed to BWESG, the obtained word vectors are again high-dimensional (typically thousands of dimensions) sparse real-valued vectors. In addition, traditional-PPMI is a purely local distributional model deriving distributional context knowledge from narrow context windows (typically 3-10 surrounding words, e.g., \citeR{Laroche:2010}). A bootstrapping approach \cite{Vulic:2013emnlp} which we use to induce the Traditional-PPMI representation starts from an automatically learned seed lexicon of one-to-one translation pairs obtained using some other model (e.g., Basic-MuPTM or Association-MuPTM), and then gradually detects new dimensions of the shared bilingual semantic space. We refer the interested reader to the relevant literature \cite{Vulic:2013emnlp} for more details. 

\subsection{Group II: BWE Induction Models Adjusted to Document-Aligned Data}
\label{ss:brm2}
\noindent {\bf BiCVM} \citeA{Hermann:2014} introduced a model called BiCVM (Bilingual Compositional Vector Model) that learns bilingual word embeddings from a sentence-aligned parallel corpus $\mathcal{C} = \{s_1,s_2,\ldots,s_N\}=\{(s_1^S,s_1^T),(s_2^S,s_2^T),\allowbreak \ldots,(s_N^S,s_N^T)\}$.\footnote{A very similar (but more expensive) model which also learns from parallel sentence-aligned data was also introduced by \citeA{Chandar:2014}.} $s_j=(s_j^S,s_j^T)$ now denotes a pair of aligned sentences. The model assumes that the aligned sentences have the same meaning, which implies that their sentence representations should be similar. Assume two functions $f$ and $g$ which map sentences given in the source and language respectively to their semantic representations in $\mathbb{R}^{d}$, where $d$ is again the representation dimensionality. The energy of the model given two sentences $(s_j^S,s_j^T) \in \mathcal{C}$ is then defined as: $E(s_j^S,s_j^T)=||f(s_j^S)-g(s_j^T)||$. The goal is to minimize $E$ for all semantically equivalent sentences (i.e., aligned sentences) in the corpus. In order to prevent the model from degenerating, they use a noise-contrastive large-margin update which ensures that the representations of non-aligned sentences observe a certain margin from each other. For every pair of parallel sentences $(s_j^S,s_j^T)$, they sample a number of additional negative sentence pairs $(s_j^S, n_{neg}^T)$ from the corpus (i.e., the sampled pairs are not observed as positive pairs in $\mathcal{C}$). These noise samples are used in formulating the hinge loss as follows: $E(s_j^S,s_j^T) = \max (mrg + \Delta E(s_j^S, s_j^T, n_{neg}^T),0)$, where $mrg$ is the margin, and $\Delta E(s_j^S, s_j^T, n_{neg}^T) = E(s_j^S,s_j^T)-E(s_j^S,n_{neg}^T)$. The loss is minimized for every pair of parallel sentences in the corpus with $L2$-regularization on the model parameters. The number of noise samples per each positive pair is a hyper-parameter of the model. A semantic signal is propagated from aligned sentences back to the individual words to obtain bilingual word embeddings. While the BiCVM model was originally built for sentence-aligned parallel data, exactly the same idea may be applied to document-aligned non-parallel data. In this paper, we test its ability to learn from noisier comparable data. The BWESG model is compared against BiCVM when inducing BWEs from both data types: comparable and parallel. \\
\\
\noindent {\bf Mikolov} Another collection of BWE induction models \cite{Mikolov:2013arxiv,Faruqui:2014,Dinu:2015arxiv,Lazaridou:2015acl} assumes the following setup: first, two monolingual embedding spaces, $\mathbb{R}^{dim_S}$ and $\mathbb{R}^{dim_T}$, are induced separately in each of the two languages using a standard monolingual WE model such as SGNS \cite{Mikolov:2013iclr,Mikolov:2013nips}. $dim_S$ and $dim_T$ denote the dimensionality of monolingual embedding spaces in the source and target language respectively. The bilingual signal is provided in the form of word translation pairs $(x_i,y_i)$, where $x_i \in V^S$, $y_i \in V^T$, and $\vec{x_i} \in \mathbb{R}^{dim_S}$, $\vec{y_i} \in \mathbb{R}^{dim_T}$. Training is cast as a multivariate regression problem: it implies learning a function that maps the source language vectors from the training data to their corresponding target language vectors. A standard approach \cite{Mikolov:2013arxiv,Dinu:2015arxiv} is to assume a linear map $\mathbf{W} \in \mathbb{R}^{dim_S \times dim_T}$, where a $L_2$-regularized least-squares error objective (i.e., ridge regression) is used to learn the map $\mathbf{W}$: it is learned by solving the following optimization problem (typically by stochastic gradient descent): \\
$\min_{\mathbf{W} \in \mathbb{R}^{dim_S \times dim_T}} ||\mathbf{XW} - \mathbf{Y}||^2_{F} + \lambda ||\mathbf{W}||^2_{F}$.

$\mathbf{X}$ and $\mathbf{Y}$ are matrices obtained through the respective concatenation of source language and target language vectors from training pairs. Once the linear map $\mathbf{W}$ is estimated, any previously unseen source language word vector $\vec{x_u}$ may be straightforwardly mapped into the target language embedding space $\mathbb{R}^{dim_T}$ as $\mathbf{W}\vec{x_u}$. After mapping all vectors $\vec{x}$, $x \in V^S$, the target embedding space $\mathbb{R}^{dim_T}$ in fact serves as a bilingual embedding space (Figure~\ref{fig:monomulti}).

Although the main strength of the model is its ability to learn embeddings on larger monolingual training sets, the model may also be adjusted to the setting where the only training data are document-aligned comparable data as follows: (1) Automatically learn a seed lexicon or reliable one-to-one translation pairs from document-aligned data using a bootstrapping approach from \citeA{Peirsman:2010,Vulic:2013emnlp}, (2) Train two separate monolingual embedding spaces on two separated halves of the document-aligned data set (i.e., using only source language documents and only target language documents), (3) Learn the mapping between the two spaces using the pairs from Step 1. \\
\\
\noindent {\bf BilBOWA} Another collection of BWE induction models {jointly} optimizes {two monolingual objectives}, with the cross-lingual objective acting as a cross-lingual regularizer during training \cite{Klementiev:2012,Gouws:2015icml,Soyer:2015iclr}. The idea behind joint training may be summarized by the simplified formulation \cite{Luong:2015nw}:
$\gamma(\text{\em Mono}_S + \text{\em Mono}_T) + \delta \text{\em Bi}$.

The monolingual objectives $Mono_S$ and $Mono_T$ ensure that similar words in each language are assigned similar embeddings and aim to capture the semantic structure of each language, whereas the cross-lingual objective $Bi$ ensures that similar words across languages are assigned similar embeddings, and ties the two monolingual spaces together into a bilingual space. Parameters $\gamma$ and $\delta$ govern the influence of the monolingual and bilingual components.\footnote{Setting $\gamma=0$ reduces the model to the setting similar to BiCVM \cite{Hermann:2014}. $\gamma=1$ results in the models from \cite{Klementiev:2012,Gouws:2015icml,Soyer:2015iclr}.} The bilingual signal for these models, now acting as the cross-lingual regularizer during the joint training, is provided in sentence-aligned parallel data. Although they use the same data sources, the models differ in the choice of monolingual and cross-lingual objectives. In this work, we opt for the BilBOWA model from \cite{Gouws:2015icml} as the representative model to be included in the comparisons, due to its previous solid performance and robustness in the BLE task, its reduced complexity reflected in fast computations on massive datasets, as well as its public availability: {\footnotesize\texttt{https://github.com/gouwsmeister/bilbowa}}. In short, the BilBOWA model combines SGNS for the monolingual objectives together with the cross-lingual objective that minimizes the $L_2$-loss between the bag-of-word vectors of parallel sentences. For more details about the exact training procedure, we refer the interested reader to the work from \citeA{Gouws:2015icml}.

Again, although the main strength of the model is its ability to learn embeddings on larger monolingual training sets, the model may also be adjusted to the setting with document- or sentence-aligned data by: (1) using two halves of the aligned corpus for separate monolingual training, (2) using the alignment signal for bilingual training. 
\section{From Word Representations to Semantic Word Similarity}
\label{s:ssimilarity}
Assume now that we have induced bilingual word representations, regardless of the chosen RM. Given two words $w_i$ and $w_j$, irrespective to their actual language, we may compute the degree of their semantic similarity by applying a {\em similarity function} (SF) on their vector representations $\overrightarrow{w_i}$ and $\overrightarrow{w_j}$: $sim(w_i,w_j)=SF(\overrightarrow{w_i},\overrightarrow{w_j})$. Different choices (or rather families of) SFs are cosine, the Kullback-Leibler or the Jensen-Shannon divergence, the Hellinger distance, the Jaccard index, etc. \cite{Lee:1999,Cha:2007}, and different RMs typically require different SFs to produce optimal or near-optimal results over various semantic tasks. When working with word embeddings, a standard choice for SF is cosine similarity ({\em cos}) \cite{Mikolov:2013nips}, which is also a typical choice in traditional distributional models \cite{Bullinaria:2007}. The similarity is then computed as follows:
\begin{align}
sim(w_i,w_j)=cos(w_i,w_j)=\frac{\overrightarrow{w_i}\cdot \overrightarrow{w_j}}{|\overrightarrow{w_i}|\cdot |\overrightarrow{w_j}|}
\end{align}
On the other hand, a good choice for SF when working with probabilistic RMs such as Basic-MuPTM and Association-MuPTM RS is the Hellinger distance \cite{Pollard:2001,Cha:2007,Kazama:2010}, which displays excellent results in the BLE task \cite{Vulic:2013naacl}. The similarity between words $w_i$ and $w_j$ using the Hellinger distance is computed as follows:
\begin{align} 
sim(w_i,w_j)=\frac{1}{\sqrt{2}}\sqrt{\sum_{i=1}^K \Big(\sqrt{P(f'_k|w_i)}-\sqrt{P(f'_k|w_j)}\Big)^2}
\end{align}
Note that the Hellinger distance is applicable only if word representations are probability distributions, which is the case for Basic-MuPTM and Association-MuPTM. $P(f'_k|w_i)$ denotes the probability score for the $k$-th dimension ($f'_k$) in the vector representation with Basic-MuPTM or Association-MuPTM.\footnote{Prior work has shown that the results for Basic-MuPTM and Association-MuPTM are slightly higher when cosine is replaced with the Hellinger distance. Therefore, in this particular case we have opted for the Hellinger distance to report a more competitive baseline.}

For each word $w_i$, we can build a {\em ranked list} $RL(w_i)$ which consists of all other words $w_j$ ranked according to their respective semantic similarity scores $sim(w_i,w_j)$. Additionally, we label the ranked list $RL(w_i)$ that is pruned at position $M$ as $RL_M(w_i)$. Since we may retain language labels for words when training in multilingual settings (e.g., language labels are marked by different colors in Figure~\ref{fig:model}), we may compute: (1) {\em monolingual similarity}, e.g., given $w_i \in V^S$, we retain only $w_j \in V^S$ in the ranked list (analogous for $w_i \in V^T)$, (2) {\em cross-lingual similarity} (CLSS), e.g., given $w_i \in V^S$, we retain only $w_j \in V^T$, and (3) {\em multilingual similarity}, where we retain all words $w_j \in V^S \cup V^T$. When computing CLSS for $w_i$, the most similar word cross-lingually is called the cross-lingual {\em nearest neighbor}. 


We will employ the models of context-insensitive CLSS at the word type level to extract bilingual lexicons from document-aligned or sentence-aligned data, and to compare all representation models in the BLE task in Section~\ref{s:ble}.

\subsection{Context Sensitive Models of (Cross-Lingual) Semantic Similarity}
\label{ss:clsscs}
The context-insensitive models of semantic similarity provide ranked lists of semantically similar words {\em invariably} or {\em in isolation}, and they operate at the level of word types. They do not explicitly encode different word senses. In practice, it means that, given a sentence {\em ``The coach of his team was not satisfied with the game yesterday.''}, these context-insensitive CLSS models are not able to detect that the Spanish word {\em entrenador} is more similar to the polysemous English word {\em coach} in the context of this sentence than the Spanish word {\em autocar}, although {\em autocar} is listed as the most semantically similar word to {\em coach} globally/invariably without any observed context. In another example, while the Spanish words {\em partido}, {\em encuentro}, {\em cerilla} or {\em correspondencia} are all highly similar to another ambiguous English word {\em match} when observed in isolation, given the Spanish sentence {\em ''She was unable to find a match in her pocket to light up a cigarette.''}, it is clear that the strength of cross-lingual semantic similarity should change in context as only {\em cerilla} exhibits a strong cross-lingual semantic similarity to {\em match} within this particular sentential context.

The goal now is to build BWE-based models of cross-lingual semantic similarity in context, similar to context-aware CLSS models proposed by \citeA{Vulic:2014emnlp}. Two key questions are: (i) How to provide BWE-based representations beyond word level to represent the context of a word token?; (ii) How to use the contextual knowledge in a context-sensitive model of semantic similarity?

Following \citeA{Vulic:2014emnlp}, given a word token $w$ in context (e.g., a window of words, a sentence, a paragraph, or a document), we build its context set or rather context bag $Con(w)=\{cw_1,\allowbreak\ldots\allowbreak,cw_r\}$ by harvesting $r$ neighboring words in the chosen context scope (e.g., the context bag may comprise all content-bearing words in the same sentence as the pivot word token, the so-called {\em sentential context}). In order to present the context $Con(w)$ in the $d$-dimensional embedding space, we need to apply a model of {\em semantic composition} to learn its $d$-dimensional vector representation $\overrightarrow{Con(w)}$. 

Formally, given word $w$, we may specify the vector representation of the context bag $Con(w)$ as the $d$-dimensional vector/embedding: 
\begin{align}
\overrightarrow{Con(w)}=\overrightarrow{cw_1} \star \overrightarrow{cw_2} \star \ldots \star \overrightarrow{cw_r}
\label{eq:composition}
\end{align}
where $\overrightarrow{cw_1},\ldots,\overrightarrow{cw_r}$ are $d$-dimensional WEs learned from the data, and $\star$ is a compositional vector operator such as addition, point-wise multiplication, tensor product, etc.

A plethora of models for semantic composition have been proposed in the relevant literature, differing in their choice of vector operators, input structures and required knowledge \cite{Mitchell:2008,BaroniZamparelli:2010,Rudolph:2010,Socher:2012,Blacoe:2012,Clarke:2012,Hermann:2014,Milajevs:2014}, to name only a few. In this work, driven by the observed linear linguistic regularities in the embedding spaces \cite{Mikolov:2013naacl}, we opt for simple {\em addition} (denoted by +) from \citeA{Mitchell:2008} as the compositional operator, due to its simplicity, the ease of applicability on bag-of-words contexts, and its relatively solid performance in various compositional tasks \cite{Mitchell:2008,Milajevs:2014}. The $d$-dimensional embedding $\overrightarrow{Con(w)}$ is then:
\begin{align}
\overrightarrow{Con(w)}=\overrightarrow{cw_1} + \overrightarrow{cw_2} + \ldots + \overrightarrow{cw_r}
\label{eq:addcomp1}
\end{align}
If we use any BWE-based RM, we may compute the context-sensitive semantic similarity score $sim(w_i,t_j,Con(w_i))$ between $t_j$ and $w_i$ given its context $Con(w_i)$ in the shared bilingual embedding space as follows:
\begin{align}
sim(w_i,t_j,Con(w_i)) = SF(\overrightarrow{w'_i},\overrightarrow{t_j})
\label{eq:addcomp2}
\end{align}
$t_j \in V^T$ is any target language word, and $\overrightarrow{t_j}$ its word representation, while $\overrightarrow{w'_i}$ is the new ``contextualized'' vector representation for $w_i$ modulated by its context $Con(w_i)$, that is, its context-aware representation. \citeA{Vulic:2014emnlp} introduced a linear interpolation of two $d$-dimensional vectors as a plausible solution for the modulation/contextualization. The modulation of representation for $w_i$ is computed as follows:
\begin{align}
\overrightarrow{w'_i} = (1-\lambda)\cdot\overrightarrow{w_i} + \lambda \cdot \overrightarrow{Con(w_i)}
\label{eq:addcomp3}
\end{align}
where $\overrightarrow{w_i}$ is the word embedding for $w_i$ computed at the word type level, $\overrightarrow{Con(w_i)}$ is the embedding for the context bag computed using eq.~\eqref{eq:addcomp1}, and $\lambda$ is an interpolation parameter.
Another set of similar models that can yield context-sensitive similarity computations has been proposed very recently, and has displayed very competitive results regardless of its simplicity \cite{Melamud:2015ws}. Here, we present two best scoring context-sensitive models which we adapt to the bilingual setting:
\begin{align}
\text{{\sc Add-Melamud: }\hspace{0.5em}} sim(w_i,t_j,Con(w_i)) = \frac{SF(w_i,t_j) + \sum_{cw_i \in Con(w_i)} SF(cw_i,t_j)}{|Con(w_i)|+1} \notag \\
\text{{\sc Mult-Melamud: }\hspace{0.0em}} sim(w_i,t_j,Con(w_i)) = \sqrt[|Con(w_i)|+1]{SF(w_i,t_j) \cdot \hspace{-0.5em}\prod_{cw_i \in Con(w_i)} SF(cw_i,t_j)} \notag
\label{eq:melamud}
\end{align}
Note that for the Mult model one has to avoid negative values, so a simple shift to an all-positives interval is required, e.g., the shifted cosine score becomes $cos'(x,y)=\frac{cos(x,y)+1}{2}$. Unlike the models from \citeA{Vulic:2014emnlp}, these two models do not aggregate single word representations into one vector that represents the context, but compute similarity scores separately with each word from the context. For more details regarding the models, we refer the interested reader to the original paper \cite{Melamud:2015ws}.

We will employ the models of context-sensitive CLSS at the word token level to compare all representation models in the task of suggesting word translations in context in Section~\ref{s:swtc}.
%
\section{Training Setup}
\noindent {\bf Training Data.} To induce bilingual word embeddings as well as to be directly comparable with baseline representations from prior work, we use a dataset comprising a subset of comparable Wikipedia data available in three language pairs \cite{Vulic:2013emnlp,Vulic:2014emnlp}\footnote{Available online: \texttt{\footnotesize people.cs.kuleuven.be/$\sim$ivan.vulic/software/}}: (i) a collection of 13,\,696 Spanish-English Wikipedia article pairs (ES-EN), (ii) a collection of 18,\,898 Italian-English Wikipedia article pairs (IT-EN),  and (iii) a collection of 7,\,612 Dutch-English Wikipedia article pairs (NL-EN). All corpora are theme-aligned comparable corpora, that is, the aligned document pairs discuss similar themes, but are in general not direct translations of each other. To be directly comparable to prior work in the two evaluation tasks \cite{Vulic:2013emnlp,Vulic:2014emnlp}, we retain only nouns that occur at least 5 times in the corpus. Lemmatized word forms are recorded when available, and original forms otherwise. TreeTagger \cite{Schmid:1994} is used for POS tagging and lemmatization. After the preprocessing steps vocabularies comprise between 7,000 and 13,000 noun types for each language in each language pair, and the training corpora are quite small: ranging from approximately 1.5M tokens for NL-EN to 4M for ES-EN. Exactly the same training data and vocabularies are used to train all representation models in comparison (both from Group I and Group II, see Section~\ref{s:baselinemodels}). 

We also demonstrate that it is simple and straightforward to train BWESG on parallel sentence-aligned data using the same modeling principles. For that purpose, we use Europarl.v7 \cite{Koehn:2005} for all three language pairs obtained from the OPUS website \cite{Tiedemann:2012lrec}.\footnote{\texttt{http://opus.lingfil.uu.se/}} As the only preprocessing step, we retain only words occurring at least 5 times in the corpus. Each corpus contains approximately 2M parallel sentences, vocabularies are by an order of magnitude larger than from the smaller Wikipedia data (i.e., varying from 45K EN word types to 75K NL word types), and the corpora sizes are approximately 120M tokens. Data statistics of the two data sources, Wikipedia vs Europarl, are provided in Table~\ref{tab:trainingstats}. The statistics reveal the different nature of the two corpora, with significantly more variance and noise reported for the Wikipedia data.
\begin{table}[t]
\centering
\begin{center}
{\scriptsize
\begin{tabularx}{\linewidth}{l XXX XXX}
\toprule
{\bf Corpus: }& \multicolumn{3}{c}{{\bf Wikipedia}} & \multicolumn{3}{c}{\bf Europarl} \\
\cmidrule(lr){2-4} \cmidrule(lr){5-7}
{\bf Pair: } & {ES-EN} & {IT-EN} & {NL-EN} & {ES-EN} & {IT-EN} & {NL-EN} \\
\midrule
{Average length (OTHER)} & {111} & {84} & {51} & {29} & {29} & {27} \\
{Average length (EN)} & {174} & {154} & {129} & {28} & {29} & {27} \\
{Average length difference} & {127} & {125} & {102} & {3} & {4} & {4} \\
\bottomrule
\end{tabularx}
}
\end{center}
\vspace{-1.2em}
\caption{Training data statistics: Non-parallel document-aligned Wikipedia vs parallel sentence-aligned Europarl for all three language pairs. OTHER = ES, IT or NL. Lengths are measured in word tokens. Averages are rounded to the closest integer.}
\label{tab:trainingstats}
\vspace{-2.2em}
\end{table}
\\
\\
\noindent {\bf Trained BWESG Models} To test the effect of random shuffling in the {\em merge and shuffle} BWESG strategy, we have trained the BWESG model with 10 random corpora shuffles for all three training corpora. We also train BWESG with the {\em length-ratio shuffle} strategy. All parameters are set to default suggested parameters for SGNS from the \texttt{word2vec} package: stochastic gradient descent (SGD) with a linearly decreasing global learning rate of 0.025, 25 negative samples, subsampling rate $1e-4$, and 15 epochs. 

We have varied the number of dimensions $d=100,200,300$. We have also trained BWESG with $d=40$ to be directly comparable to readily available sets of BWEs from prior work \cite{Chandar:2014}. Moreover, to test the effect of window size on the final results, i.e., the number of positives used for training, we have varied the maximum window size $cs$ from 4 to 60 in steps of 4.\footnote{We remind the reader that we slightly abuse terminology here, as the BWESG windows do not include the locality component any more.}

We will make our pre-training and training code for BWESG publicly available, along with all BWESG-based bilingual word embeddings for the three language pairs at: \\ \texttt{http://liir.cs.kuleuven.be/software.php}. \\
\\
\noindent {\bf Baseline Representations: Group I } All parameters of the baseline representation models (i.e., topic models and their settings, the number of dimensions $K$, the values for feature pruning, window size, weighting and similarity functions) were optimized in prior work. Therefore, the settings are adopted directly from previous work \cite{Griffiths:2007,Bullinaria:2007,Dinu:2010,Vulic:2013naacl,Vulic:2013emnlp,Kiela:2014cvsc}, and we encourage the interested reader to check the details and exact parameter setup in the relevant literature. We provide only a short overview here.

For Basic-MuPTM and Association-MuPTM, as in \cite{Vulic:2013naacl}, a bilingual latent Dirichlet allocation (BiLDA) model was trained with $K=2000$ topics and the standard values for hyper-parameters: $\alpha = 50/K$, $\beta=0.01$ \cite{Steyvers:2007}. Post-hoc semantic space pruning was employed with the pruning parameter set to 200 for Basic-MuPTM and to 2000 for Association-MuPTM. We refer the reader to the relevant paper for more details.

For Traditional-PPMI, as in \cite{Vulic:2013emnlp}, a seed lexicon was automatically obtained by bootstrapping from the initial seed lexicon of reliable pairs stemming from the Association-MuPTM model (with the same parameters for Association-MuPTM as listed above). The window size was fixed to 6 in both directions. We again refer the reader to the paper for more details. \\
\\
\noindent {\bf Baseline Representations: Group II } All baseline BWE models were trained with the same number of dimensions as BWESG: $d=100,200,300$. Other model-specific parameters were taken as suggested in prior work.

For {\sc BiCVM}, we use the tool released by the authors.\footnote{\texttt{https://github.com/karlmoritz/bicvm}} We train an additive model, with hinge loss margin $mrg=d$ as in the original paper, batch size of 50, and noise parameter of 10. All models were trained with 200 iterations.

For {\sc Mikolov}, we train two monolingual SGNS models using the original \texttt{word2vec} package, SGD with a global learning rate of 0.025, 25 negative samples, subsampling rate $1e-4$, and 15 epochs. The seed lexicon required to learn the mapping between two monolingual spaces is exactly the same as for Traditional-PPMI.

For BilBOWA, we use SGD with a global learning rate 0.15 for training\footnote{Suggestions for parameter values received through personal correspondence with the authors. The software is available online: \texttt{https://github.com/gouwsmeister/bilbowa}}, 25 negative samples, subsampling rate $1e-4$, and 15 epochs. For BilBOWA and {\sc Mikolov}, we vary the window size the same way as in BWESG. \\
\\
\noindent {\bf Similarity Functions } Unless stated otherwise, a similarity function used in all similarity computations with all RMs is cosine ({\em cos}). The only exceptions are Basic-MuPTM and Association-MuPTM where the Hellinger distance (HD) was used since it consistently outperformed cosine for these two RM types in prior work (see Footnote~7). \\
\\
\noindent {\bf A Roadmap to Experiments } In the first experiment, we quickly visually inspect the obtained lists of semantically similar words using the BWESG bilingual representation model. Following that, we compare BWESG-based models for bilingual lexicon extraction (BLE) and suggesting word translations in context (SWTC) against both groups of baseline models discussed in Section~\ref{s:baselinemodels}. The experiments and results for the BLE task are presented in Section~\ref{s:ble}, while the experiments and results for SWTC are presented in Section~\ref{s:swtc}.
%
\section{Evaluation Task I: Bilingual Lexicon Extraction}
\label{s:ble}
\subsection{Task Description}
One may employ the context-insensitive CLSS models from Section~\ref{s:ssimilarity} to extract bilingual lexicons automatically from data. By harvesting cross-lingual nearest neighbors, one is able to build a bilingual lexicon of one-to-one translation pairs ($w_i^S,w_j^T$). We test the validity of our BWEs and baseline representations in the BLE task.

\subsection{Experimental Setup}
\label{ss:expsetup}
\noindent {\bf Test Data } For each language pair, we evaluate on standard 1,000 ground truth one-to-one translation pairs built for the three language pairs (ES/IT/NL-EN) by \citeA{Vulic:2013naacl,Vulic:2013emnlp}. Translation direction is ES/IT/NL $\rightarrow$ EN. The data is available online.\footnote{\texttt{http://people.cs.kuleuven.be/~ivan.vulic/software/}} \\
\\
\noindent {\bf Evaluation Metrics } Since we can build a one-to-one bilingual lexicon by harvesting one-to-one translation pairs, the lexicon quality is best reflected in the $Acc_1$ score, that is, the number of source language (ES/IT/NL) words $w_i^S$ from ground truth translation pairs for which the top ranked word cross-lingually is the correct translation in the other language (EN) according to the ground truth over the total number of ground truth translation pairs ({\em =1000}) \cite{Gaussier:2004,Tamura:2012,Vulic:2013emnlp}. Similar trends are observed within a more lenient setting with $Acc_{5}$ and $Acc_{10}$ scores, but we omit these results for clarity and the fact that the actual BLE performance is best reflected in $Acc_1$.\\
\begin{table}[t]
	\centering
	\scriptsize{
	\begin{tabularx}{1.0\linewidth}{XXX|XXX|XXX}
		\toprule
		\multicolumn{3}{c}{{\bf Spanish-English (ES-EN)}} & \multicolumn{3}{c}{\bf Italian-English (IT-EN)} & \multicolumn{3}{c}{\bf Dutch-English (NL-EN)} \\
		\midrule
		{(1)} & {(2)} & {(3)} & {(1)} & {(2)} & {(3)} & {(1)} & {(2)} & {(3)} \\
		{{\bf reina}} & {{\bf reina}} & {{\bf reina}} & {{\bf madre}} & {{\bf madre}} & {{\bf madre}} & {{\bf schilder}} & {{\bf schilder}} & {{\bf schilder}} \\
		\midrule
		{(Spanish)} & {(English)} & {(Combined)} & {(Italian)} & {(English)} & {(Combined)} & {(Dutch)} & {(English)} & {(Combined)} \\
		\midrule
		\midrule
		{rey} & {\em queen(+)} & {\em queen(+)} & {padre} & {\em mother(+)} & {\em mother(+)} & {kunstschilder} & {\em painter(+)} & {\em painter(+)} \\
		{trono} & {\em heir} & {rey} & {moglie} & {\em father} & {padre} & {schilderij} & {\em painting} & {kunstschilder} \\
		{monarca} & {\em throne} & {trono} & {sorella} & {\em sister} & {moglie} & {kunstenaar} & {\em portrait} & {\em painting} \\
		{heredero} & {\em king} & {\em heir} & {figlia} & {\em wife} & {\em father} & {olieverf} & {\em artist} & {schilderij} \\
		{matrimonio} & {\em royal} & {\em throne} & {figlio} & {\em daughter} & {sorella} & {portret} & {\em canvas} & {kunstenaar} \\
		{hijo} & {\em reign} & {monarca} & {fratello} & {\em son} & {figlia} & {schilderen} & {\em brush} & {\em portrait} \\
		{reino} & {\em succession} & {heredero} & {casa} & {\em friend} & {figlio} & {frans} & {\em cubism} & {olieverf} \\
		{reinado} & {\em princess} & {\em king} & {amico} & {\em childhood} & {\em sister} & {nederlands} & {\em art} & {portret} \\
		{regencia} & {\em marriage} & {matrimonio} & {marito} & {\em family} & {fratello} & {componist} & {\em poet} & {schilderen} \\
		{duque} & {\em prince} & {\em royal} & {donna} & {\em cousin} & {\em wife} & {beeldhouwer} & {\em drawing} & {\em artist} \\
		\bottomrule
	\end{tabularx}
		\caption{Example lists of top 10 semantically similar words for all 3 language pairs obtained using BWESG (length-ratio shuffle); $d=200, cs=48$; (col 1.) only source language words (ES/IT/NL) are listed while target language words are skipped (monolingual similarity); (2) only target language words (EN) are listed (cross-lingual similarity); (3) words from both languages are listed (multilingual similarity). The correct one-to-one translation is marked by (+).}
	\label{tab:ex}
\vspace{-2.2em}
}
\end{table}

%

\subsection{Results and Discussion}
\vspace{-0.0em}
\label{ss:blerd}
\subsubsection{Experiment 0: Qualitative Analysis and Comparison} 
Table~\ref{tab:ex} displays top 10 semantically similar words monolingually, across-languages and combined/multilingually for one ES, IT and NL word. BWESG is able to find semantically coherent lists of words for all three directions of similarity (i.e., monolingual, cross-lingual, multilingual). In the combined (multilingual) ranked lists, words from both languages are represented as top similar words. This initial qualitative analysis already demonstrates the ability of BWESG to induce a shared bilingual embedding space using only document alignments as bilingual signals.\footnote{We also conducted a small experiment on solving word analogies using monolingual English embedding spaces, and then we repeated the experiment with the same vocabulary and bilingual English-Spanish/Italian/Dutch embedding spaces. The results follow the findings from \cite{Faruqui:2014}, where only slight (and often insignificant) fluctuations for SGNS vectors were reported (e.g., the fluctuations are $<1$\% on average in our experiments) when moving from monolingual to bilingual embedding spaces. We may conclude that the linguistic regularities \cite{Mikolov:2013naacl} established for monolingual embedding spaces induced by SGNS also hold in bilingual embedding spaces induced by BWESG.}

In another brief analysis, we qualitatively compare the cross-lingual ranked lists acquired by BWESG with the other three baseline CLSS/BLE models from Group I. The lists for one ES word and one IT word are presented in Table~\ref{tab:comp}. For the two example words, BWESG is the only model which is able to rank the actual correct translations as nearest cross-lingual neighbors. It is already symptomatic that the word {\em gulf}, which is the correct translation for {\em golfo}, does not occur in the ranked list $RL_{10}(golfo)$ at all in case of the three baseline models. We will soon quantitatively confirm this initial suspicion, and demonstrate that BWESG is superior to the three baseline models in the BLE task.
\begin{table}[t]
	\centering
	\scriptsize{
	\begin{tabularx}{\linewidth}{XXXX|XXXX}
		\toprule
		\multicolumn{4}{c}{{\bf Spanish-English (ES-EN)}} & \multicolumn{4}{c}{\bf Italian-English (IT-EN)} \\
		\midrule
		{BWESG} & {BMu} & {AMu} & {TPPMI} & {BWESG} & {BMu} & {AMu} & {TPPMI} \\
		\midrule
		{\bf cebolla} & {\bf cebolla} & {\bf cebolla} & {\bf cebolla} & {\bf golfo} & {\bf golfo} & {\bf golfo} & {\bf golfo}\\
		\midrule
		\midrule
		{onion(+)} & {dessert} & {dessert} & {sauce} & {gulf(+)} & {whale} & {coast} & {coast} \\
		{dish} & {salad} & {walnut} & {cheese} & {coast} & {dolphin} & {isthmus} & {sea} \\
		{marinade} & {nut} & {salad} & {garlic} & {coastline} & {coast} & {coastline} & {island} \\
		{cuisine} & {walnut} & {nut} & {salad} & {bay} & {suborder} & {fjord} & {bay} \\
		{soup} & {rice} & {hazelnut} & {chili} & {island} & {cadmium} & {ferry} & {lagoon} \\
		{sauce} & {toast} & {porridge} & {onion(+)} & {peninsula} & {ferry} & {monsoon} & {harbour} \\
		{cheese} & {porridge} & {rice} & {cuisine} & {settlement} & {monsoon} & {mainland} & {beach} \\
		{coriander} & {paddy} & {marinade} & {flavor} & {shore} & {fjord} & {seaside} & {shore} \\
		{vegetable} & {tuber} & {toast} & {bread} & {tourism} & {isthmus} & {isle} & {river} \\
		{tortilla} & {potato} & {paddy} & {dish} & {ferry} & {mainland} & {suborder} & {lake} \\
		\bottomrule
	\end{tabularx}
	\vspace{-0.8em}
			\caption{Example lists of top 10 semantically similar words for ES-EN and IT-EN, obtained using BWESG (length-ratio, $d=200,cs=48$), and the three representation models from Group I. The correct translation is marked by (+).}
	\label{tab:comp}
\vspace{-2.2em}
}
\end{table}

As an aside, Table~\ref{tab:comp} also clearly reveals the difficulty of judging the quality of models for computing semantic similarity/relatedness solely based on the observed output of the models. The lists $RL_{10}(cebolla)$ and $RL_{10}(golfo)$ appear significantly different across all four models, yet all these lists contain words which appear semantically related to the source word. Therefore, we require a more systematic quantitative task-oriented comparison of induced word representations.

\subsubsection{Experiment I: BWESG vs Group I}
\label{ss:bwe1sect}
Table~\ref{tab:exp1} shows the first set of results on the BLE task: we report scores with two different BWESG strategies as well as with a BWESG model which does not shuffle pseudo-bilingual documents. The previous best reported $Acc_1$ scores with baseline representations for the same training+test combination are also reported in the table. By zooming into the table multiple times, we summarize the most important findings.
\begin{table}[t]
\centering
\begin{center}
{\scriptsize
\begin{tabularx}{\linewidth}{l XXX XXX XXX}
\toprule
{\bf Pair: }& \multicolumn{3}{c}{{\bf ES-EN}} & \multicolumn{3}{c}{\bf IT-EN} & \multicolumn{3}{c}{\bf NL-EN} \\
\cmidrule(lr){2-4} \cmidrule(lr){5-7} \cmidrule(lr){8-10}
{\sc \scriptsize \bf BWESG} & {$d$=$100$} & {$d$=$200$} & {$d$=$300$} & {$d$=$100$} & {$d$=$200$} & {$d$=$300$} & {$d$=$100$} & {$d$=$200$} & {$d$=$300$} \\
{\sc \scriptsize \bf Merge and Shuffle} & {} & {} & {} & {} & {} & {} & {} & {} & {} \\
{\scriptsize $cs$:16,MIN} & {0.607} & {0.600} & {0.577} & {0.585} & {0.597} & {0.571} & {0.293} & {0.244} & {0.219} \\
{\scriptsize $cs$:16,AVG} & {0.617} & {0.613} & {0.596} & {0.599} & {0.601} & {0.583} & {0.300} & {0.254} & {0.224} \\
{\scriptsize $cs$:16,MAX} & {0.625} & {0.630} & {0.613} & {0.607} & {0.606} & {0.596} & {0.307} & {0.267} & {0.233} \\
{\scriptsize $cs$:48,MIN} & {0.658} & {0.676} & {0.672} & {0.662} & {0.677} & {0.672} & {0.378} & {0.366} & {0.354} \\
{\scriptsize $cs$:48,AVG} & {0.665} & {0.685} & {0.688} & {0.669} & {0.683} & {0.683} & {0.389} & {0.381} & {0.363} \\
{\scriptsize $cs$:48,MAX} & {0.675} & {0.694} & {\bf 0.705} & {0.677} & {\bf 0.692} & {0.689} & {0.394} & {0.395} & {0.377} \\
\midrule
{\sc \scriptsize \bf BWESG}  & {$d$=$100$} & {$d$=$200$} & {$d$=$300$} & {$d$=$100$} & {$d$=$200$} & {$d$=$300$} & {$d$=$100$} & {$d$=$200$} & {$d$=$300$} \\
{\sc \scriptsize \bf Length-Ratio} & {} & {} & {} & {} & {} & {} & {} & {} & {} \\
{\scriptsize $cs$:16} & {0.627} & {0.610} & {0.602} & {0.613} & {0.614} & {0.595} & {0.303} & {0.275} & {0.237} \\
{\scriptsize $cs$:48} & {\bf 0.678} & {\bf 0.701} & {0.703} & {\bf 0.679} & {0.689} & {\bf 0.692} & {\bf 0.397} & {\bf 0.396} & {\bf 0.382} \\
\midrule
{\sc \scriptsize \bf BWESG}  & {$d$=$100$} & {$d$=$200$} & {$d$=$300$} & {$d$=$100$} & {$d$=$200$} & {$d$=$300$} & {$d$=$100$} & {$d$=$200$} & {$d$=$300$} \\
{\bf \scriptsize No Shuffling} & {} & {} & {} & {} & {} & {} & {} & {} & {} \\
{\scriptsize $cs$:16} & {0.218} & {0.176} & {0.139} & {0.209} & {0.198} & {0.162} & {0.070} & {0.068} & {0.049} \\
{\scriptsize $cs$:48} & {0.511} & {0.497} & {0.480} & {0.523} & {0.540} & {0.526} & {0.214} & {0.198} & {0.197} \\
\midrule
\midrule
{\sc \scriptsize \bf BMu} & {0.441} & {0.441} & {0.441} & {0.575} & {0.575} & {0.575} & {0.237} & {0.237} & {0.237} \\
{\sc \scriptsize \bf AMu} & {0.518} & {0.518} & {0.518} & {0.618} & {0.618} & {0.618} & {0.236} & {0.236} & {0.236} \\
{\sc \scriptsize \bf TPPMI} & {0.577} & {0.577} & {0.577} & {0.647} & {0.647} & {0.647} & {0.206} & {0.206} & {0.206} \\
\bottomrule
\end{tabularx}
}
\end{center}
\vspace{-1.0em}
\caption{BLE performance in terms of $Acc_1$ scores for all tested BLE models for Spanish-English, Italian-English and Dutch-English with all bilingual word representations learned from document-aligned Wikipedia data. For BWESG with {\em merge and shuffle} we report maximum (MAX), minimum (MIN) and average (AVG) scores over 10 random corpora shuffles. Highest scores per column are in bold.}
\label{tab:exp1}
\vspace{-1.0em}
\end{table} 

\noindent {\bf BWESG vs Baseline Representations } The results clearly reveal the superior performance of the BWESG model for BLE which relies on our new framework for inducing bilingual word embeddings from document-aligned comparable data over other BLE models relying on previously used bilingual word representations from the same type of training data. The increase in $Acc_1$ scores over the best scoring baseline models is 22.2\% for ES-EN, 7\% for IT-EN and 67.5\% for NL-EN. 

\noindent {\bf BWESG Shuffling Strategy } Although both BWESG strategies display results that are above established baselines, there is a clear advantage to the {\em length-ratio shuffle} strategy, which displays a solid and robust performance across a variety of parameters and all three language pairs. Another advantage of that strategy is the fact that it has a deterministic outcome and does not suffer from ``sub-optimal'' random shuffles. In summary, we suggest using the {\em length-ratio shuffle} strategy in future work, and along the same line we opt for that strategy in all further experiments.

\begin{figure}[t]
           \centering
			\subfigure[Spanish-English]{
                                \includegraphics[scale=0.37]{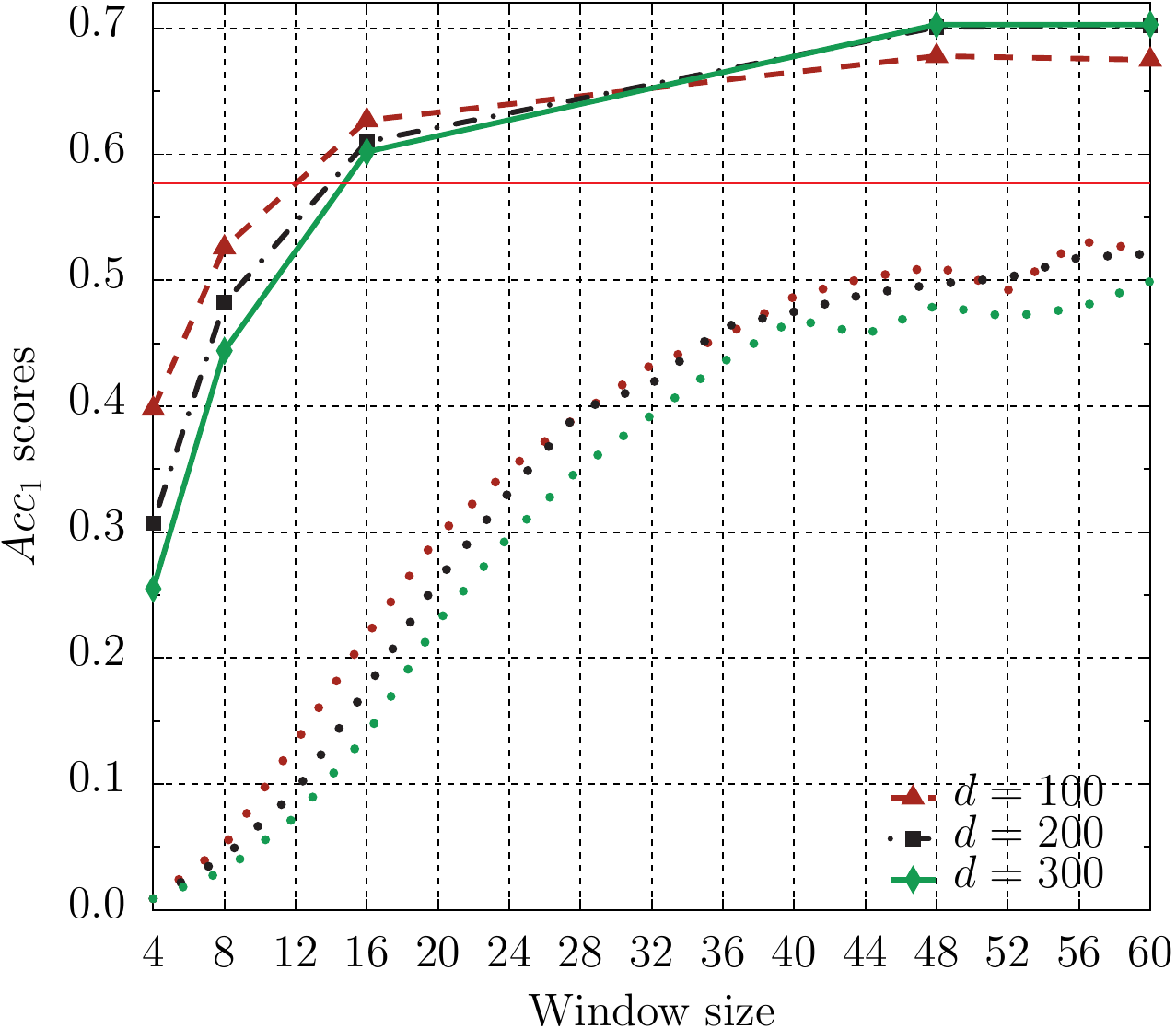}
				\label{fig:dim100jair}
			}
           \subfigure[Italian-English]{
                                \includegraphics[scale=0.37]{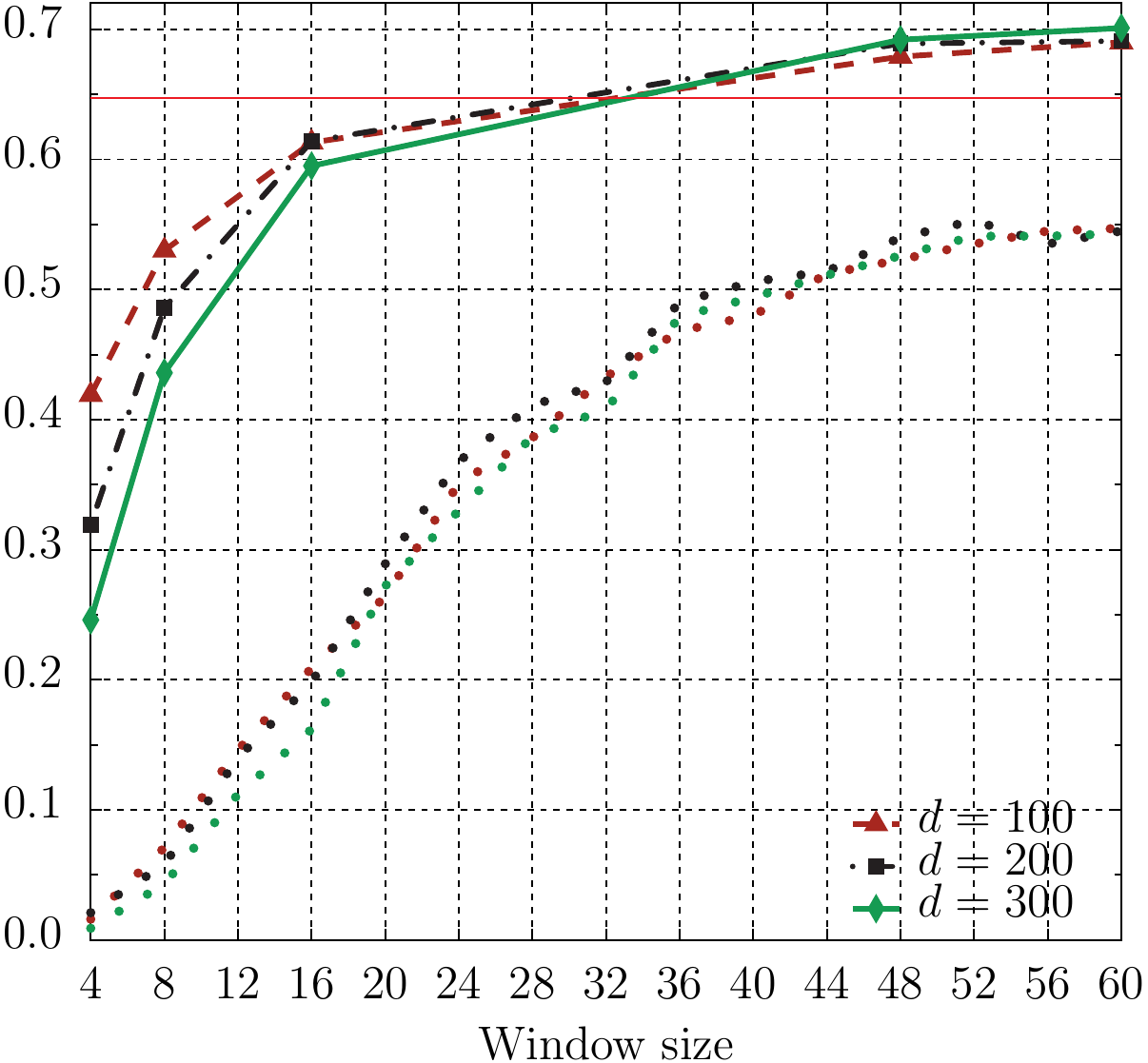}
				\label{fig:dim200jair}
			}
			\subfigure[Dutch-English]{
                                \includegraphics[scale=0.37]{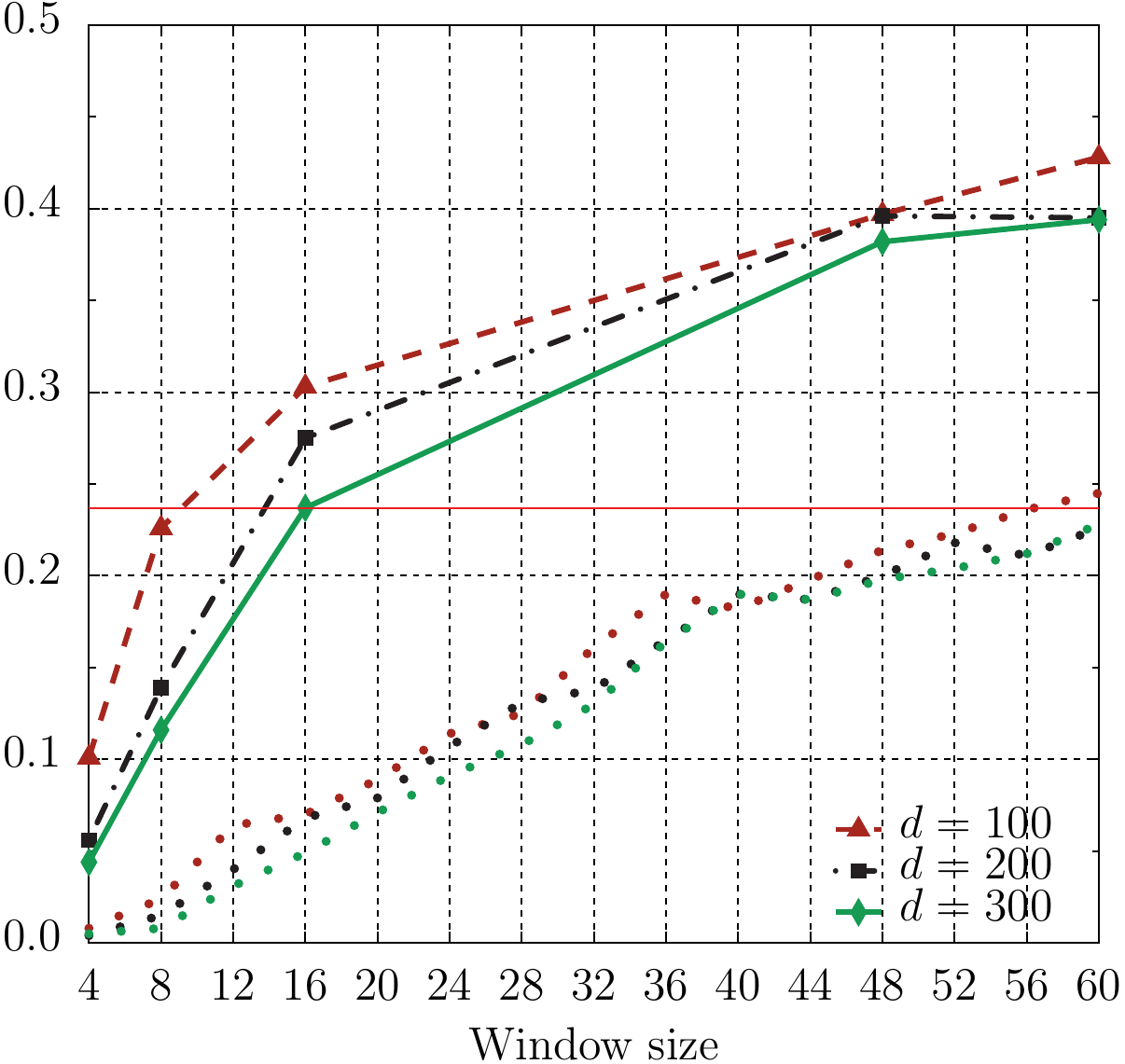}
				\label{fig:dim300jair}
			}
			\vspace{-1.0em}			
			\caption{$Acc_1$ scores in the BLE task with BWESG {\em length-ratio shuffle} for all 3 language pairs, and varying values for parameters $cs$ and $d$. Solid (red) horizontal lines denote the highest baseline $Acc_1$ scores for each language pair. Thicker dotted lines refer to BWESG without shuffling.}
			\vspace{-1.8em}
		 \end{figure}

The results also reveal that shuffling is universally useful, as BWESG without shuffling relies largely on monolingual contexts and cannot reach the performance of BWESG with shuffling. A partial remedy for the problem is to train BWESG with more document-level training pairs (i.e., by increasing the window size), but that leads to prohibitively expensive models, and nonetheless BWESG without shuffling with larger $cs$-s still falls short of BWESG with both shuffling strategies (see also Figures~\ref{fig:dim100jair}-\ref{fig:dim300jair}). 

\noindent {\bf Window Size: Number of Training Pairs } The results confirm the intuition that larger window sizes, i.e., more training examples lead to better results in the BLE task. For all embedding dimensions $d$-s, BWESG exhibits a superior performance for $cs=48$ than for $cs=16$, and the performance with $cs=48$ and $cs=60$ seems relatively stable: intuitively, more training pairs leads to a slightly better BLE performance, but the curve slowly flattens out (Figures~\ref{fig:dim100jair}-\ref{fig:dim300jair}). This finding reveals that even a coarse tuning of these parameters might lead to optimal or near-optimal scores for BLE with BWESG. 

\noindent {\bf Differences across Language Pairs } A lower increase in $Acc_1$ scores for IT-EN is attributed to the fact that the test set for IT-EN comprises IT words with occurrence frequencies above 200 in the training data \cite{Vulic:2013naacl}, while the other two test sets comprise randomly sampled words covering all frequency spectra. As expected, all models in comparison are able to effectively utilize distributional signals for higher-frequency words, but BWESG still displays the best performance, and these improvements in $Acc_1$ scores are statistically significant (using McNemar's statistical significance test, $p<0.05$).\footnote{McNemar's significance test is very common in the NLP literature, especially when $Acc_1$ scores are reported. It utilizes the standard 2$\times$2 contingency table, and may be observed as a paired version of the more common chi-square test. The reader is referred to the original work \cite{McNemar:1947}.}

Further, the lowest overall scores for all models in comparison are observed for NL-EN. We attribute it to using less training data for NL-EN when compared to ES-EN and IT-EN (i.e., training corpora for ES-EN and IT-EN are almost triple the size of training corpora for NL-EN). However, we observe that the increase obtained by BWESG is even more prominent in this setting with limited training data.  The lower results of TPPMI compared to other two baseline models are also attributed to the overall lower quality and size of NL-EN training data, which is then reflected in a lower quality of seed lexicons necessary to start the bootstrapping procedure from \citeA{Vulic:2013emnlp}. 

\noindent {\bf Computational Complexity } BWESG trained with larger values for $d$ and $cs$ yields richer semantic representations, but also naturally leads to increased training times. However, due to a lightweight design of the supporting SGNS, the times are by the order of magnitude lower than the training times for Basic-MuPTM or Association-MuPTM. Typically, several hours are needed to train BWESG with $d=300$ and $cs\approx 48-60$, whereas it takes two to three days to train a bilingual topic model with $K=2000$ on the same training set using the multi-threaded architectures on 10 Intel(R) Xeon(R) CPU E5-2667 2.90GHz processors. The BWESG model scales as expected (i.e., training time increases linearly with the window size with all other parameters being equal), and enjoys all the advantages (training time-wise and memory-wise) of the original \texttt{word2vec} package. A logical explanation for the behaviour follows from the interpretation of SGNS provided by \citeA{Levy:2014acl}, e.g., using a window size of 48 instead of a window size 16 basically means using 3 times more positive examples for training (e.g., approximately 15 minutes is needed to train $300$-dimensional ES-EN BWESG embeddings with $cs=16$ using the Wikipedia data as opposed to 46 minutes with $cs=48$, measured again on 10 Intel(R) Xeon(R) processors).

\subsubsection{Experiment II: BWESG vs Other BWE Induction Models (Group II)}
\label{ss:exp2ble}
\begin{table}[t]
\centering
\begin{center}
{\scriptsize
\begin{tabularx}{\linewidth}{l XXX XXX XXX}
\toprule
{\bf Pair: }& \multicolumn{3}{c}{{\bf ES-EN}} & \multicolumn{3}{c}{\bf IT-EN} & \multicolumn{3}{c}{\bf NL-EN} \\
\cmidrule(lr){2-4} \cmidrule(lr){5-7} \cmidrule(lr){8-10}
{\sc \scriptsize \bf BWESG} & {$d$=$100$} & {$d$=$200$} & {$d$=$300$} & {$d$=$100$} & {$d$=$200$} & {$d$=$300$} & {$d$=$100$} & {$d$=$200$} & {$d$=$300$} \\
{\sc \scriptsize \bf Length-Ratio} & {} & {} & {} & {} & {} & {} & {} & {} & {} \\
{\scriptsize $cs$:48} & {\bf 0.678} & {\bf 0.701} & {\bf 0.703} & {\bf 0.679} & {\bf 0.689} & {\bf 0.692} & {\bf 0.397} & {\bf \bf 0.396} & {\bf 0.382} \\
\midrule
{\scriptsize {\sc \bf Mikolov}} & {} & {} & {} & {} & {} & {} & {} & {} & {} \\
{\scriptsize $cs$:4} & {0.187} & {0.151} & {0.282} & {0.368} & {0.382} & {0.533} & {0.042} & {0.068} & {0.120} \\
{\scriptsize $cs$:8} & {0.305} & {0.306} & {0.420} & {0.462} & {0.518} & {0.582} & {0.076} & {0.095} & {0.145} \\
{\scriptsize $cs$:16} & {0.344} & {0.396} & {0.486} & {0.472} & {0.539} & {0.602} & {0.117} & {0.161} & {0.184} \\
{\scriptsize $cs$:48} & {0.311} & {0.375} & {0.477} & {0.458} & {0.536} & {0.591} & {0.132} & {0.178} & {0.202} \\
{\scriptsize $cs$:60} & {0.324} & {0.389} & {0.479} & {0.460} & {0.538} & {0.597} & {0.151} & {0.180} & {0.209} \\
\midrule
{\scriptsize {\sc \bf BiCVM}} & {} & {} & {} & {} & {} & {} & {} & {} & {} \\
{iterations:200} & {0.342} & {0.384} & {0.403} & {0.309} & {0.366} & {0.377} & {0.068} & {0.084} & {0.083} \\
\bottomrule
\end{tabularx}
}
\end{center}
\vspace{-1.5em}
\caption{BLE results: Comparison of BWESG with (1) the BWE induction model from \citeA{Mikolov:2013arxiv} relying on SGNS, (2) BiCVM: the BWE induction model from \citeA{Hermann:2014} initially developed for parallel sentence-aligned data. All models were trained on the same document-aligned training Wikipedia data with exactly the same vocabularies.}
\label{tab:expmikolov}
\vspace{-1.2em}
\end{table} 
All further experiments are conducted using BWESG with the {\em length-ratio shuffle} strategy. Note that again all models in comparison use exactly the same data sources and vocabularies as BWESG and Group I models from the previous section. The results with BiCVM and the {\sc Mikolov} model are summarized in Table~\ref{tab:expmikolov}: the comparison reveals a clear and prominent advantage for the BWESG model given the same data and training setup. We do not report absolute scores of the BilBOWA model in this setup as they were much lower than the other two baseline models. The BiCVM model, although in theory fit to learn from document-aligned data, is unable to compete with BWESG when learning BWEs from the noisier setting with non-parallel data.

\label{ss:exp3parallel}
\begin{figure}[t]
           \centering
			\subfigure[Spanish-English]{
                                \includegraphics[scale=0.41]{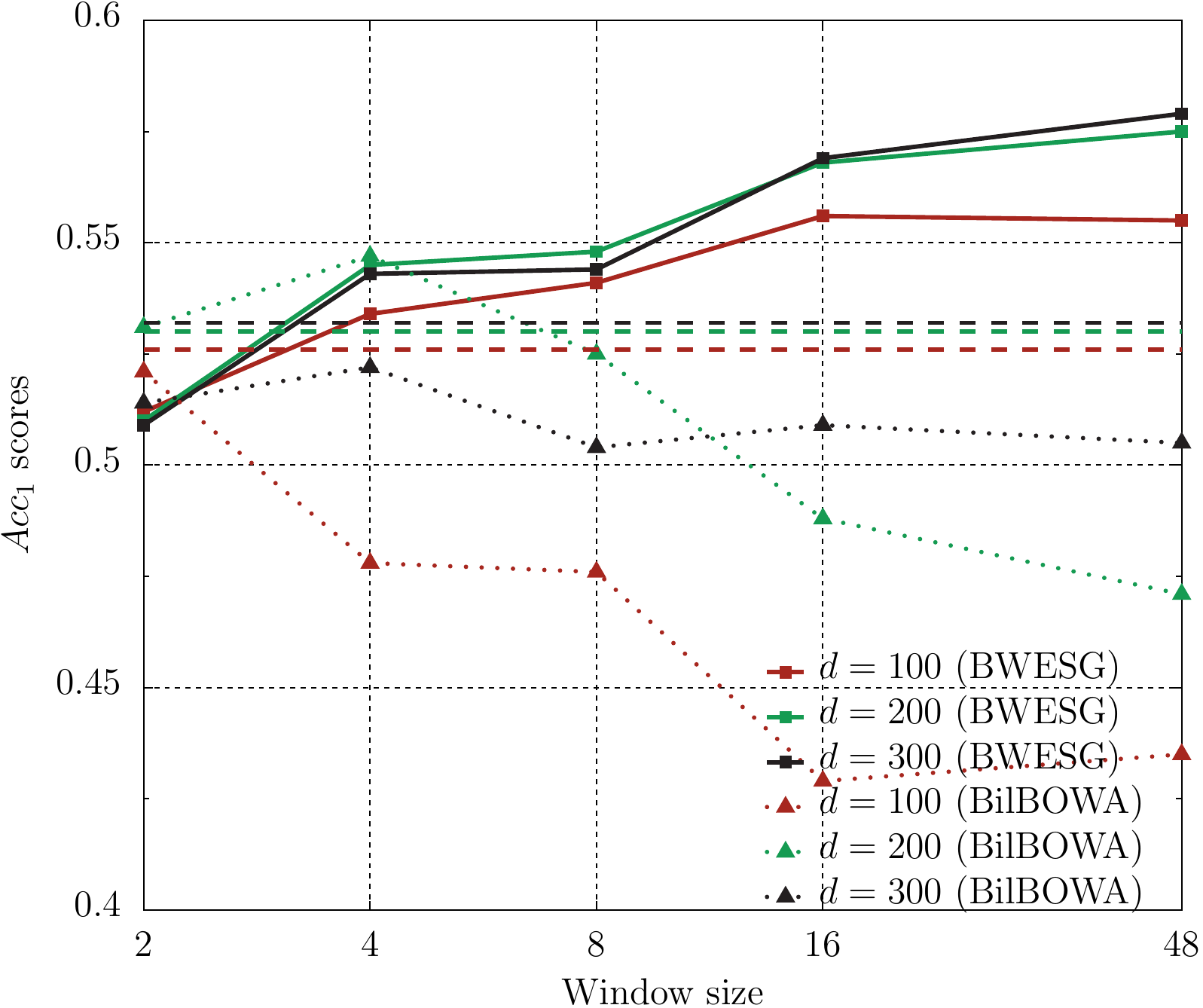}
				\label{fig:dim100ep}
			}
           \subfigure[Italian-English]{
                                \includegraphics[scale=0.41]{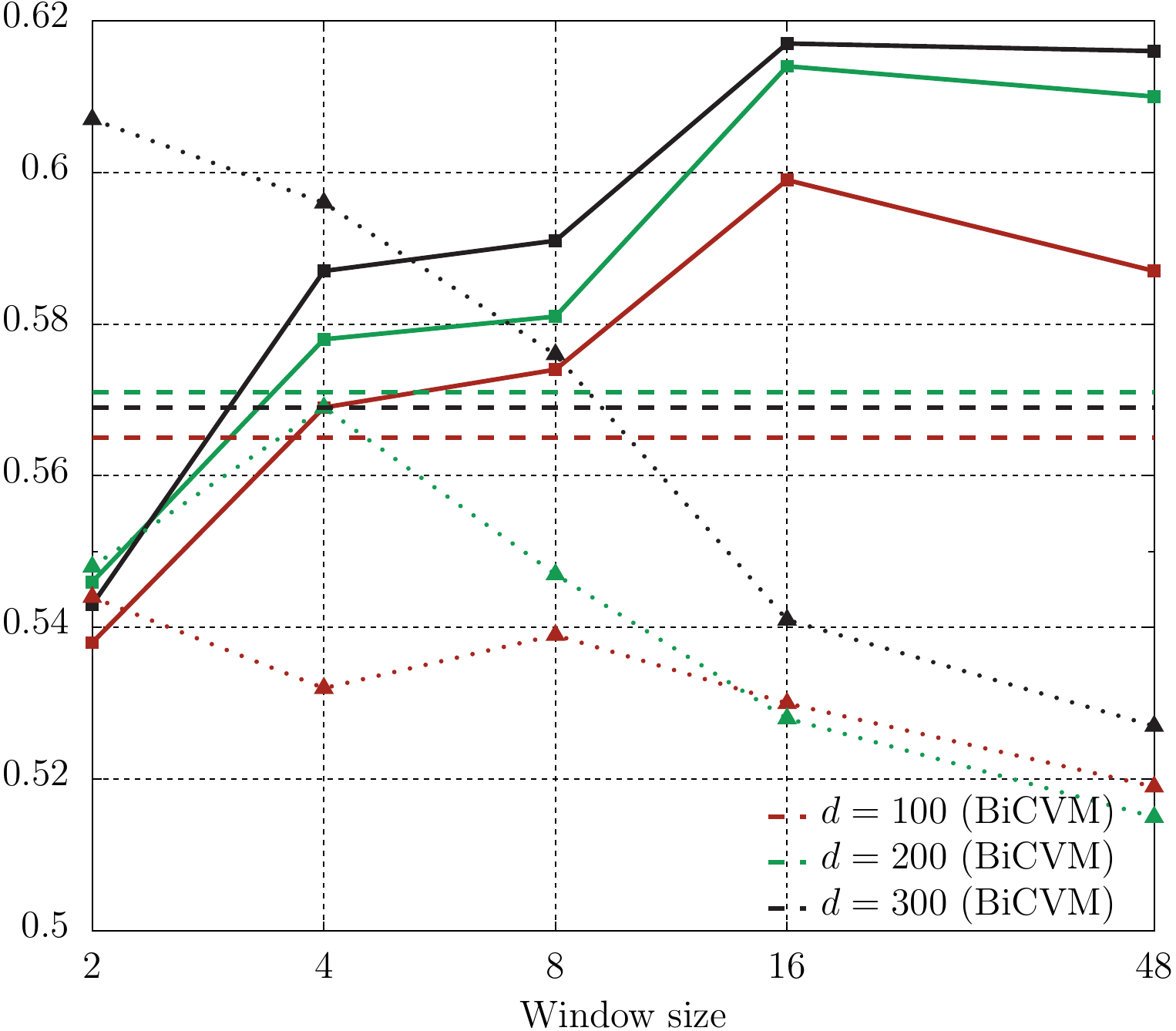}
				\label{fig:dim200ep}
			}
			\subfigure[Dutch-English]{
                                \includegraphics[scale=0.41]{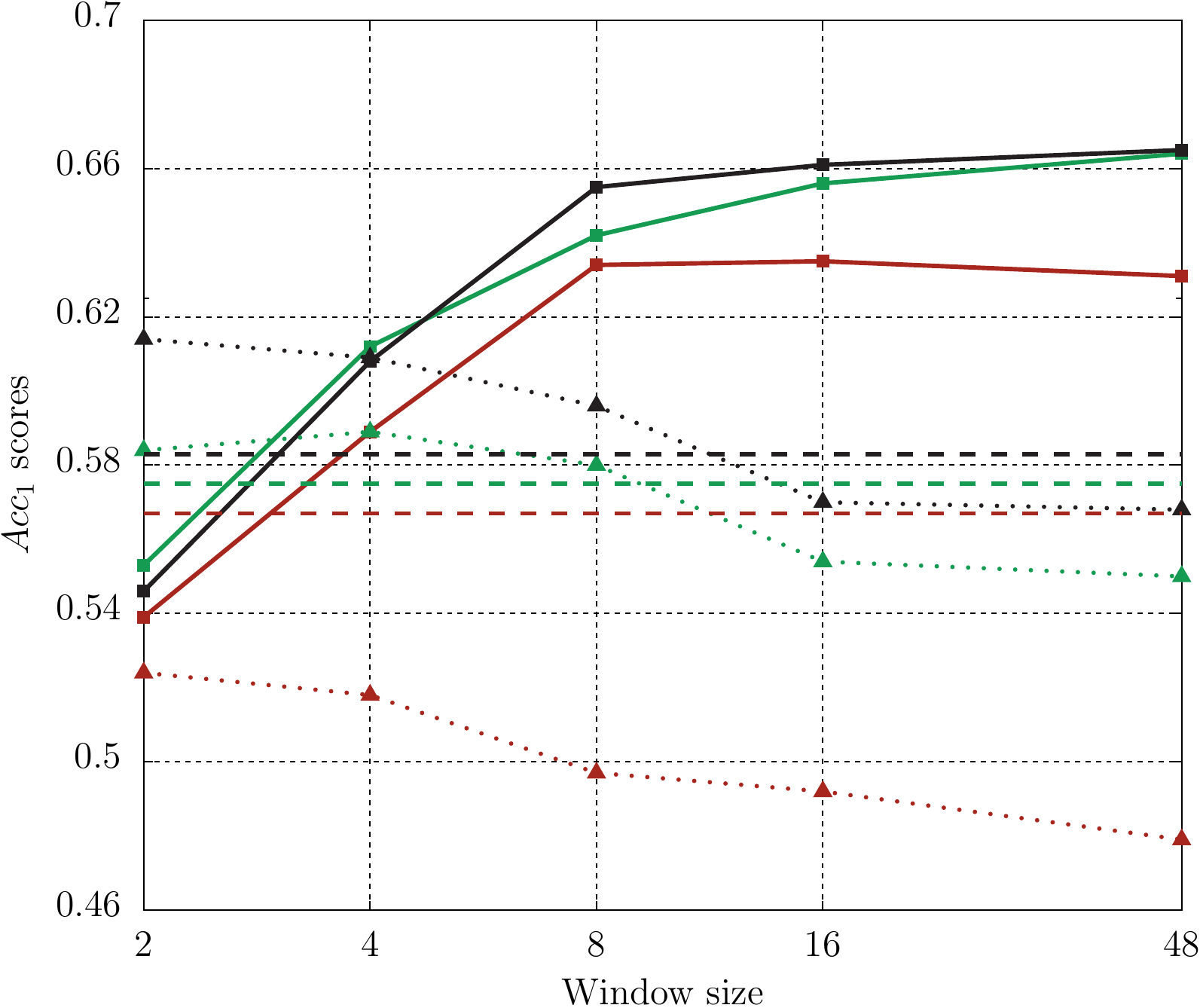}
				\label{fig:dim300ep}
			}
			\vspace{-1.0em}			
			\caption{Comparison of BWESG (solid curves) with two other models that rely on parallel training data: (1) BilBOWA (dotted curves), (2) BiCVM: the BWE induction modelinitially developed for parallel sentence-aligned data (dashed horizontal lines). All models were trained on the same sentence-aligned training Europarl data with exactly the same vocabularies. BLE is performed over the same search space for all models. $x$ axes are in log scale.}
			\vspace{-1.5em}
		 \end{figure}
We also present a preliminary study where we compare BWSESG and Group II models in the setup with parallel sentence-aligned data. Results are summarized in Figures~\ref{fig:dim100ep}-\ref{fig:dim300ep}.\footnote{Note that the absolute scores are not directly comparable to the BLE scores when the model is trained on Wikipedia data (Tables~\ref{tab:exp1} and \ref{tab:expmikolov}) due to different training data, different preprocessing steps and vocabularies. Different vocabularies also result in different BLE search spaces and coverages of the test sets (e.g., some very common Spanish nouns from the test set such as {\em nadador (swimmer)} or {\em colmillo (tusk)} are not observed in Europarl due to the domain shift).} The preliminary results clearly demonstrate that BWESG is able to learn BWEs from parallel data without the slightest change in its modeling principles. While the BilBOWA model displays better results for lower values of the $cs$ parameter, to our own surprise, the BWESG model is comparable to or even better than the baseline models with larger window sizes. The BiCVM model, which implicitly utilizes the entire sentence span in training also outperforms BWESG with smaller windows, but BWESG again performs significantly better with larger windows. The BWESG performance flattens out quicker than with the Wikipedia data (compare the results with $cs=16$ and $cs=48$), which is easily explained by the decreased length of aligned items as provided in Table~\ref{tab:trainingstats} (i.e., sentences vs documents).

For English-Spanish, we can also compare BWESG to pre-trained $40$-dimensional embeddings from \citeA{Chandar:2014}, as their embeddings were also induced on the same Europarl data. While their model's $Acc_1$ score is 0.432 for $d=40$, BWESG obtains $Acc_1$ scores of 0.502 ($d=40$, $cs=8$), 0.535 ($d=40$, $cs=16$) or 0.529 ($d=40$, $cs=48$).
%
\section{Evaluation Task II: Suggesting Word Translations in Context}
\label{s:swtc}
In another task, we test the ability of BWEs to produce context-sensitive semantic similarity modeling (see Section~\ref{ss:clsscs}), which in turn may be used to solve the task of suggesting word translations in context (SWTC) proposed recently \cite{Vulic:2014emnlp}. The goal now is to build BWESG-based models for SWTC given the sentential context, similar as in the prior work. We show that our new BWESG-based SWTC models outperform the best SWTC models \cite{Vulic:2014emnlp}, as well as other SWTC models which rely on the baseline word representations discussed in Section~\ref{s:baselinemodels}.

\subsection{Task Description}
Given an occurrence of a polysemous word $w_i \in V^S$ and the context of that occurrence, the SWTC task is to choose the correct translation in the target language $L_T$ of that particular occurrence of $w_i$ from the given set $\mathcal{TC}(w_i)=\{t_1,\ldots,t_{tq}\}$, $\mathcal{TC}(w_i) \subseteq V^T$, of its $tq$ possible translations/meanings. We may refer to $\mathcal{TC}(w_i)$ as an {\em inventory of translation candidates} for $w_i$. The task of {\em suggesting word translations in context} (SWTC) may be interpreted as ranking the $tq$ translation candidates with respect to the observed local context $Con(w_i)$ of the occurrence of the word $w_i$. The best scoring translation candidate according to the scores $sim(w_i,t_j,Con(w_i))$ (see Section~\ref{ss:clsscs}) in the ranked list is then the correct translation for that particular occurrence of $w_i$ observing its local context $Con(w_i)$. 

\subsection{Experimental Setup}
\label{ss:swtcexp}
\noindent {\bf Test Data } We use the SWTC test set introduced recently \cite{Vulic:2014emnlp}. The test set comprises 15 polysemous nouns in three languages (ES, IT and NL) along with sets of their translation candidates (i.e., sets $\mathcal{TC}$). For each polysemous noun, the test sets provide 24 sentences extracted from Wikipedia which illustrate different senses and translations of the pivot polysemous noun, accompanied by the annotated correct translation for each sentence. It yields 360 test sentences for each language pair (and 1080 test sentences in total). An additional set of 100 IT sentences (5 other polysemous IT nouns plus 20 sentences for each noun) is used as a development set to tune the parameter $\lambda$ (see Section~\ref{ss:clsscs}) for all language pairs and all models in comparison. In summary, the final aim may be formulated as follows: For each polysemous word $w_i$ in ES/IT/NL, the goal is to suggest its correct translation in English given its sentential context. \\
\\
\noindent {\bf Evaluation Metrics } Since the task is to present a list of possible translations to a SWTC model, and then let the model decide a single most likely translation given the word and its sentential context, we measure the performance again as {\em Top 1} accuracy ($Acc_1$). 

\subsection{Results and Discussion}
\label{ss:swtcrd}
\subsubsection{Experiment I: BWESG vs Group I}
Note that the Group I models held previously best reported SWTC scores for the training+test combination.\\
\\
\noindent {\bf Models in Comparison } (1) {\em BWESG+add}. RM: BWESG. SF: cos. Composition: addition. $\lambda=1.0$. The value for $\lambda$ suggests that only context is used to disambiguate the meaning of a polysemous word and to guess its most likely translation in context.\footnote{We have also experimented with the context-sensitive CLSS models proposed by \citeA{Melamud:2015ws}, but we do not report the actual scores as this model, although displaying a similar relative ranking of different representation models, was consistently outperformed by the models from \citeA{Vulic:2014emnlp} in our evaluation runs: $\approx$0.75-0.80 vs $\approx$0.60-0.65 for the models from \citeA{Melamud:2015ws}.}\\
(2) {\em BMu+HD+S}. RM: BasicMuPTM. SF: Hellinger distance. Composition: Smoothed-Fusion\footnote{In short, Smoothed-Fusion is a probabilistic variant of the context-sensitive modeling idea presented by equations~\eqref{eq:addcomp1}-\eqref{eq:addcomp3}. For more details, check \cite{Vulic:2014emnlp}.} from \cite{Vulic:2014emnlp}. $\lambda=0.9$. \\
(3) {\em BMu+Cue+S}. RM: BasicMuPTM. SF: Cue or Association measure \cite{Steyvers:2007,Vulic:2013naacl}. Composition: Smoothed-Fusion. $\lambda=0.9$. The Cue similarity is tailored for probabilistic models and computed as the association score $P(t_i|w_i')=\sum_{k=1}^K P(t_i|z_k)P(z_k|w_i')$, where $z_k$ denotes $k$-th latent feature, and $P(z_k|w_i')$ denotes the modulated probability score obtained by smoothing the probabilistic representations of $w_i$ and its context $Con(w_i)$. \\
(4) {\em TPPMI+add}. RM: Traditional-PPMI. SF: cos. Composition: addition. $\lambda=0.9$. 

Again, all parameters of the baseline representation models are adopted directly from prior work where they were optimized on development sets comprising additional 100 sentences \cite{Vulic:2014emnlp}. In addition, BMu+HD+S and BMu+Cue+S also rely on the procedure of context sorting and pruning \cite{Vulic:2014emnlp}, where the idea is to retain only context words which are most semantically similar to the given pivot polysemous word, and then use them in computations. The procedure, however, produces significant gains only for probabilistic models (BMu+HD+S and BMu+Cue+S), and therefore, we employ it only for these models. BMu+HD+S and BMu+Cue+S with context sorting and pruning were the best scoring models in the introductory SWTC paper \cite{Vulic:2014emnlp} and currently produce state-of-the-art SWTC results on these test sets.\footnote{We omit results for the Association-MuPTM RM since SWTC models based on Association-MuPTM were consistently outperformed by SWTC models based on Basic-MuPTM across different settings.}

Table~\ref{tab:expswtc} summarizes the results and comparison with Group I models on the SWTC task. NO-CONTEXT refers to the context-insensitive majority baseline (i.e., always choosing the most semantically similar translation candidate obtained by BWESG at the word type level, without taking into account any context information).
\begin{table}[t]
\centering
\begin{center}
{\scriptsize
\begin{tabularx}{\linewidth}{l XXX XXX XXX}
\toprule
{\bf Pair: }& \multicolumn{3}{c}{{\bf ES-EN}} & \multicolumn{3}{c}{\bf IT-EN} & \multicolumn{3}{c}{\bf NL-EN} \\
\cmidrule(lr){2-4} \cmidrule(lr){5-7} \cmidrule(lr){8-10}
{\bf \scriptsize BWESG+add} & {$d$=$100$} & {$d$=$200$} & {$d$=$300$} & {$d$=$100$} & {$d$=$200$} & {$d$=$300$} & {$d$=$100$} & {$d$=$200$} & {$d$=$300$} \\
{\bf \scriptsize Length-Ratio} & {} & {} & {} & {} & {} & {} & {} & {} & {} \\
{\scriptsize $cs$:16} & {\bf 0.794*} & {\bf 0.767*} & {0.752*} & {\bf 0.817*} & {0.789} & {0.794} & {0.778*} & {0.769*} & {0.767*} \\
{\scriptsize $cs$:48} & {0.752*} & {0.758*} & {\bf 0.764*} & {0.814*} & {\bf 0.831*} & {\bf 0.814*} & {\bf 0.797*} & {\bf 0.789*} & {\bf 0.775*} \\
\midrule
{\bf \scriptsize BWESG+add}  & {$d$=$100$} & {$d$=$200$} & {$d$=$300$} & {$d$=$100$} & {$d$=$200$} & {$d$=$300$} & {$d$=$100$} & {$d$=$200$} & {$d$=$300$} \\
{\bf \scriptsize No Shuffling} & {} & {} & {} & {} & {} & {} & {} & {} & {} \\
{\scriptsize $cs$:16} & {0.717} & {0.717} & {0.694} & {0.747} & {0.728} & {0.728} & {0.722} & {0.686} & {0.678} \\
{\scriptsize $cs$:48} & {0.731} & {0.692} & {0.686} & {0.775} & {0.778} & {0.758} & {0.739} & {0.733} & {0.719} \\
\midrule
\midrule
{\bf \scriptsize NO-CONTEXT} & {0.406} & {0.406} & {0.406} & {0.408} & {0.408} & {0.408} & {0.433} & {0.433} & {0.433} \\
\midrule
{\bf \scriptsize BMu+HD+S} & {0.664} & {0.664} & {0.664} & {0.731} & {0.731} & {0.731} & {0.669} & {0.669} & {0.669} \\
{\bf \scriptsize BMu+Cue+S} & {0.703} & {0.703} & {0.703} & {0.761} & {0.761} & {0.761} & {0.712} & {0.712} & {0.712} \\
{\bf \scriptsize TPPMI+add} & {0.619} & {0.619} & {0.619} & {0.706} & {0.706} & {0.706} & {0.614} & {0.614} & {0.614} \\
\bottomrule
\end{tabularx}
}
\end{center}
\vspace{-1.0em}
\caption{A comparison of SWTC models for Spanish-English, Italian-English and Dutch-English with all bilingual word representations learned from document-aligned Wikipedia data. The asterisk (*) denotes statistically significant improvements of BWESG+add over the strongest baseline according to a McNemar's statistical significance test ($p<0.05$). Highest scores per column are in bold.}
\label{tab:expswtc}
\vspace{-2.5em}
\end{table} 
\begin{table}[b]
\centering
\begin{center}
{\footnotesize
\begin{tabularx}{1.01\linewidth}{X l l l}
\toprule
{\bf Senses: }& \multicolumn{1}{l}{{\bf 2 senses}} & \multicolumn{1}{l}{\bf 3 senses} & \multicolumn{1}{l}{\bf 4 senses} \\
\cmidrule(lr){2-2} \cmidrule(lr){3-3} \cmidrule(lr){4-4}
{\bf Model} & {$Acc_1$} & {$Acc_1$} & {$Acc_1$} \\
\toprule
{\footnotesize BMu+Cue+S} & {0.827} & {0.619} & {0.417}\\
{\footnotesize BWESG+add} & {\bf 0.834} & {\bf 0.804} & {\bf 0.583}\\
\bottomrule
\end{tabularx}
}
\end{center}
\vspace{-1.2em}
\caption{A comparison of the best scoring baseline model BMu+Cue+S and the best scoring BWESG+add model over different clusters of words (2-sense, 3-sense and 4-sense words) for Spanish-English.}
\label{tab:exp2swtc}
\vspace{-2.0em}
\end{table}

\noindent {\bf BWESG vs Baseline Representations } The results reveal that BWESG outperforms baseline bilingual word representations from Group I also in the SWTC task. The improvements are prominent for all reported values of parameters $d$ and $cs$, and are often statistically significant even when compared to the strongest baseline (which is the fine-tuned BMu+Cue+S model with context sorting and pruning for all three language pairs from \citeA{Vulic:2014emnlp}). The increase in $Acc_1$ scores over the strongest baseline is 12.9\% for ES-EN, 11.9\% for IT-EN, and 12.4\% for NL-EN. The obtained results surpass previous state-of-the-art scores and are currently the best reported results on the SWTC datasets when using non-parallel data to learn semantic representations. \\
\noindent {\bf BWESG Shuffling Strategy } Although BWESG without shuffling (due to a reduced complexity of the SWTC task compared to BLE) already displays encouraging results, there is again a clear advantage to the {\em length-ratio shuffle} strategy, which displays an excellent performance for all three language pairs. In simple words, shuffling is again useful. \\
\noindent {\bf Dimensionality and Number of Training Pairs } Unlike in the BLE task, the highest $Acc_1$ scores on average are obtained by using lower-dimensional word embeddings (i.e., $d=100$). The phenomenon may be attributed to the effect of semantic composition and the reduced complexity of the SWTC task compared to the BLE task. First, although enlarging the dimensionality of embeddings leads to an increased semantic expressiveness within the shared bilingual embedding space, it may be harmful when working with composition models, since the simple additive model of semantic composition may produce more erroneous dimensions when constructing higher-dimensional context embeddings out of single word embeddings. Second, due to its design, the SWTC task requires coarser-grained representations than BLE. While in the BLE task the goal is to detect a translation of a word from a vocabulary which typically spans (tens of) thousands of words, in the SWTC task the goal is to detect the most likely translation of a word given its sentential context, but from a small closed vocabulary of 2-4 possible translations from the translation inventory. Therefore, it is highly likely that even low-dimensional embeddings are sufficient to produce plausible rankings for the SWTC task, while at the same time, they are not sufficient and expressive enough to find correct translations in BLE. More training pairs (i.e., larger windows) still yield better results on average in the SWTC task. In summary, the choice of representation granularity is dependent on the actual task, which consequently leads to the conclusion that optimal values for $d$ and $cs$ are largely task-specific (compare also results in Table~\ref{tab:exp1} and Table~\ref{tab:expswtc}). \\
\noindent {\bf Testing Polysemy } In order to test whether the gain in performance for BWESG+add is derived mostly from the effective handling of the easiest set of words, that is, bisemous words (polysemous words with only 2 translation candidates), we have performed an additional experiment, where we have measured $Acc_1$ scores separately for words with 2, 3, and 4 different senses. Results indicate that the performance gain comes mostly from gains on trisemous and tetrasemous words, while the scores on bisemous words are comparable. Table~\ref{tab:exp2swtc} shows $Acc_1$ over different clusters of words for ES-EN, and similar scoring patterns are observed for IT-EN and NL-EN. \\
\noindent {\bf Differences across Language Pairs } Due to the reduced complexity of SWTC, we may also observe relatively higher results for NL-EN when compared to ES-EN and IT-EN, as opposed to their relative performance in the BLE task, where the scores for NL-EN are much lower than scores for ES-EN and IT-EN. Since SWTC is a less difficult task which requires coarse-grained representations, even limited amounts of training data may be sufficient to learn word embeddings which are useful for the specific task. This finding is in line with the recent work from \citeA{Gouws:2015naacl}.

\subsubsection{Experiment II: BWESG vs. Other BWE Induction Models (Group II)}
\begin{table}[t]
\centering
\begin{center}
{\scriptsize
\begin{tabularx}{\linewidth}{l XXX XXX XXX}
\toprule
{\bf Pair: }& \multicolumn{3}{c}{{\bf ES-EN}} & \multicolumn{3}{c}{\bf IT-EN} & \multicolumn{3}{c}{\bf NL-EN} \\
\cmidrule(lr){2-4} \cmidrule(lr){5-7} \cmidrule(lr){8-10}
{\sc \scriptsize \bf BWESG+add} & {$d$=$100$} & {$d$=$200$} & {$d$=$300$} & {$d$=$100$} & {$d$=$200$} & {$d$=$300$} & {$d$=$100$} & {$d$=$200$} & {$d$=$300$} \\
{\sc \scriptsize \bf Length-Ratio} & {} & {} & {} & {} & {} & {} & {} & {} & {} \\
{\scriptsize $cs$:16} & {\bf 0.794} & {\bf 0.767} & {0.752} & {\bf 0.817} & {0.789} & {0.794} & {0.778} & {0.769} & {0.767} \\
{\scriptsize $cs$:48} & {0.752} & {0.758} & {\bf 0.764} & {0.814} & {\bf 0.831} & {\bf 0.814} & {\bf 0.797} & {\bf 0.789} & {\bf 0.775} \\
\midrule
{\scriptsize {\sc Mikolov}} & {} & {} & {} & {} & {} & {} & {} & {} & {} \\
\midrule
{\scriptsize $cs$:4} & {0.742} & {0.739} & {0.725} & {0.733} & {0.706} & {0.692} & {0.692} & {0.700} & {0.700} \\
{\scriptsize $cs$:8} & {0.767} & {0.750} & {0.747} & {0.767} & {0.747} & {0.744} & {0.694} & {0.697} & {0.672} \\
{\scriptsize $cs$:16} & {0.769} & {0.744} & {0.747} & {0.758} & {0.755} & {0.758} & {0.725} & {0.700} & {0.689} \\
{\scriptsize $cs$:48} & {0.678} & {0.642} & {0.669} & {0.714} & {0.714} & {0.747} & {0.725} & {0.711} & {0.708} \\
{\scriptsize $cs$:60} & {0.636} & {0.658} & {0.656} & {0.725} & {0.725} & {0.742} & {0.722} & {0.728} & {0.722} \\
\midrule
{\scriptsize {\sc \bf BiCVM}} & {} & {} & {} & {} & {} & {} & {} & {} & {} \\
{iterations:200} & {0.547} & {0.567} & {0.539} & {0.636} & {0.664} & {0.642} & {0.586} & {0.567} & {0.581} \\
\bottomrule
\end{tabularx}
}
\end{center}
\vspace{-1.5em}
\caption{SWTC results: Comparison of BWESG with (1) the BWE induction model from \citeA{Mikolov:2013arxiv} relying on SGNS, (2) BiCVM: the BWE induction model from \citeA{Hermann:2014} initially developed for parallel sentence-aligned data. All models were trained on the same document-aligned training Wikipedia data with exactly the same vocabularies.}
\label{tab:swtcmikolov}
\vspace{-2.2em}
\end{table}  
We again test other BWE induction models in the SWTC task, using the same training setup and sets of embeddings as introduced in Section~\ref{ss:exp2ble} for the BLE task. The representations were now plugged in the context-sensitive CLSS modeling framework from Section~\ref{ss:clsscs}, and the optimization of parameters for SWTC has been conducted in the same manner as for BWESG. The results with the {\sc Mikolov} model and BiCVM are summarized in Table~\ref{tab:swtcmikolov}. The results with BilBOWA are very similar to BiCVM, so we do not report it for brevity.

BWESG outperforms other BWE induction models in the SWTC task and further confirms its utility in cross-lingual semantic modeling. The model from \citeA{Mikolov:2013arxiv} constitutes a stronger baseline: Good results in the SWTC task with this model are an interesting finding per se. While the model is not competitive with BWESG and other baseline representations models from document-aligned data in a more difficult BLE task when using noisy one-to-one translation pairs, its performance on the less complex SWTC task with a reduced search space is solid even when the model relies on the imperfect set of translation pairs to learn the mapping between two monolingual embedding spaces.


\subsubsection{Further Discussion} 


By analyzing the influence of pre-training shuffling on the results in two different evaluation tasks, we may safely establish its utility when inducing bilingual word embeddings using the BWESG model. While we have already presented two shuffling strategies in this work, one line of future work will investigate different possibilities of ``blending in'' words from two different vocabularies into pseudo-bilingual documents in a more structured and systematic manner. For instance, one approach to generating pseudo-training sentences for learning from textual and perceptual modalities has been recently introduced \cite{Hill:2014emnlp}. However, it is not straightforward how to extend this approach to the generation of pseudo-bilingual training documents. 

Another idea in the same vein is to build artificial training data of higher-quality starting from noisy comparable data by: (1) computing semantically similar words monolingually and across-languages from the noisy data, (2) retaining only highly reliable pairs of similar words using an automatic selection procedure \cite{Vulic:20122}, (3) building pseudo-bilingual documents using only reliable context word pairs. In other words, the questions is: Is it possible to choose positive training pairs more systematically to reduce the noise stemming from non-parallel data? The construction of such artificial training data and training on such data would then proceed in a bootstrapping fashion, and the model should be able to steadily reduce noise inherently present in comparable data. The idea of ``improving corpus comparability'' was only touched upon in previous work \cite{Li:2010,Li:2011}.

While the entire framework proposed in this article is in theory completely language pair agnostic as it does not make any language pair dependent modeling assumptions, we acknowledge the fact that all three language pairs comprise languages coming from the same phylum, that is, the Indo-European language family. Future extensions also include porting the framework to other more distant language pairs that do not share the same roots nor the same alphabet (e.g., English-Chinese/Hindi/Arabic), and for which benchmarking test sets are still scarce for a variety of semantic tasks (e.g., SWTC) \cite{Camacho:2015acl}. We believe that larger window sizes may solve difficulties with different word orderings (e.g., for Chinese-English). 
%
\section{Conclusions and Future Work}
\label{s:conclusions}
We have proposed and described Bilingual Word Embeddings Skip-Gram (BWESG), a simple yet effective bilingual word representation learning model which is able to induce bilingual word embeddings solely on the basis of document-aligned comparable data. BWESG is based on the omnipresent skip-gram with negative sampling (SGNS). We have presented two ways to build pseudo-bilingual documents on which a monolingual SGNS (or any monolingual WE induction model) may be trained to produce shared bilingual embedding spaces. The BWESG model does not make any language-pair dependent assumptions nor requires language-pair specific external resources such as bilingual lexicons, predefined category/ontology knowledge or parallel data. We have showed that the model may be trained on non-parallel and parallel data without any changes in modeling principles, which, complemented with its simplicity and lightweight design makes it potentially very useful as a tool for researchers in machine translation and information retrieval. 

We have employed induced BWEs in two semantic tasks: (1) bilingual lexicon extraction (BLE), and (2) suggesting word translations in context (SWTC). Our new BWESG-based BLE and SWTC models outperform previous state-of-the-art models for BLE and SWTC from document-aligned comparable data and related BWE induction models \cite{Mikolov:2013arxiv,Chandar:2014,Gouws:2015icml}. The findings in this article follow the recently published surveys from \citeA{Baroni:2014,Levy:2015tacl} regarding a solid and robust performance of neural word representations/word embeddings in semantic tasks: our new BWESG-based models for BLE and SWTC significantly outscore previous state-of-the-art distributional approaches on both tasks across different parameter settings. Even more encouraging is the fact that these new state-of-the-art results are attained using default parameter settings for the BWESG model as suggested in the \texttt{word2vec} package without any development set. Further (finer) tuning of model parameters in future work may lead to higher-quality bilingual embedding spaces.

Several straightforward lines of future research have already been tackled in Sections~\ref{s:ble} and ~\ref{s:swtc}. For instance, the current {\em length-ratio} shuffling strategy may be replaced by a more advanced shuffling method in future work. Moreover, BWEs induced by BWESG may be used in other semantic tasks besides the ones discussed in this work, and it would be interesting to experiment with other types of context aggregation and selection beyond the bag-of-words assumption, such as dependency-based contexts \cite{Levy:2014acl}, or other objective functions during training in the same vein as proposed by \citeA{Levy:2014}. Similar to the evolution in multilingual probabilistic topic modeling, another path of future work may lead to investigating bilingual models for learning BWEs which will be able to jointly learn from separate documents in aligned document pairs, without the need to construct pseudo-bilingual documents.

A natural step in the text representation learning research is to extend the focus from single word representations to composite phrase, sentence and document representations \cite{Hermann:2013,Kalchbrenner:2014,Le:2014,Soyer:2015iclr}. In this article, we have relied on a simple composition model based on vector addition, and have shown that this model performs excellent in the SWTC task. However, in the long run this model is not by any means sufficient to effectively capture all complex compositional phenomena in the data. Several models which aim to learn sentence and document embeddings have been proposed recently, but they critically rely on sentence-aligned parallel data. It is yet to be seen how to build structured multilingual phrase, sentence and document embeddings solely on the basis of comparable data. Such low-cost multilingual embeddings beyond the word level extracted from comparable data may find its application in a variety of tasks such as statistical machine translation \cite{Mikolov:2013arxiv,Zou:2013,Zhang:2014,Wu:2014}, semantic tasks such as multilingual semantic textual similarity \cite{Agirre:2014semeval}, cross-lingual information retrieval \cite{Vulic:20121,Vulic:2015sigir} or cross-lingual document classification \cite{Klementiev:2012,Hermann:2014,Chandar:2014}. 

In another future research path, we may use the knowledge of BWEs obtained by BWESG from document-aligned data to learn bilingual correspondences (e.g., word translation pairs or lists of semantically similar words across languages) which may in turn be used for learning from large unaligned multilingual datasets \cite{Mikolov:2013arxiv,AlRfou:2013}. In the long run, this idea may lead to large-scale learning models from huge amounts of multilingual data without any requirement for parallel data or manually built bilingual lexicons.




\vskip 0.2in
\bibliography{jair2015_main_r2}
\bibliographystyle{theapa}

\end{document}